\documentclass[10pt,twocolumn,letterpaper]{article}

\usepackage[pagenumbers]{cvpr} 

\newcommand{\myparagraph}[1]{{\noindent\textbf{#1}}}

\usepackage{bbm}
\usepackage{float}

\usepackage{microtype}

\definecolor{cvprblue}{rgb}{0.21,0.49,0.74}
\usepackage[breaklinks,colorlinks,allcolors=cvprblue]{hyperref}
\usepackage{graphicx}
\usepackage{booktabs}
\usepackage{siunitx}
\usepackage[table]{xcolor}
\usepackage{multirow}

\def\paperID{44358} 
\def\confName{CVPR}
\def\confYear{2026}

\newcommand{\methodname}{EZPC}

\title{Explaining CLIP Zero-shot Predictions Through Concepts}

\author{
Onat Ozdemir\textsuperscript{1,2} \enspace
Anders Christensen\textsuperscript{3,4,5} \enspace
Stephan Alaniz\textsuperscript{6} \enspace
Zeynep Akata\textsuperscript{7,8,9,10} \enspace
Emre Akbas\textsuperscript{2,8,11} \\[0.5em]
\textsuperscript{1}School of Informatics, University of Edinburgh \\
\textsuperscript{2}Dept. of Computer Eng., Middle East Technical University (METU) \quad
\textsuperscript{3}Orbital \\
\textsuperscript{4}DTU Compute, Technical University of Denmark \quad
\textsuperscript{5}Dept. of Biology, University of Copenhagen  \\
\textsuperscript{6}LTCI, Télécom Paris, Institut Polytechnique de Paris \quad
\textsuperscript{7}Technical University of Munich (TUM) \\
\textsuperscript{8}Helmholtz Munich \quad
\textsuperscript{9}MCML \quad
\textsuperscript{10}MDSI \quad
\textsuperscript{11}Robotics \& AI Center (ROMER), METU \\[0.5em]
}

\begin{document}
\maketitle

\begin{abstract}
Large-scale vision-language models such as CLIP have achieved remarkable success in zero-shot image recognition, yet their predictions remain largely opaque to human understanding. In contrast, Concept Bottleneck Models provide interpretable intermediate representations by reasoning through human-defined concepts, but they rely on concept supervision and lack the ability to generalize to unseen classes.  We introduce EZPC that bridges these two paradigms by explaining CLIP’s zero-shot predictions through human-understandable concepts. Our method projects CLIP’s joint image-text embeddings into a concept space learned from language descriptions, enabling faithful and transparent explanations without additional supervision. The model learns this projection via a combination of alignment and reconstruction objectives, ensuring that concept activations preserve CLIP’s semantic structure while remaining interpretable. Extensive experiments on five benchmark datasets, CIFAR-100, CUB-200-2011, Places365, ImageNet-100, and ImageNet-1k, demonstrate that our approach maintains CLIP’s strong zero-shot classification accuracy while providing meaningful concept-level explanations. By grounding open-vocabulary predictions in explicit semantic concepts, our method offers a principled step toward interpretable and trustworthy vision-language models. Code is available at \href{https://github.com/oonat/ezpc}{https://github.com/oonat/ezpc}.
\end{abstract}

\section{Introduction}
\begin{figure*}[t]
    \centering
    \includegraphics[width=\linewidth]{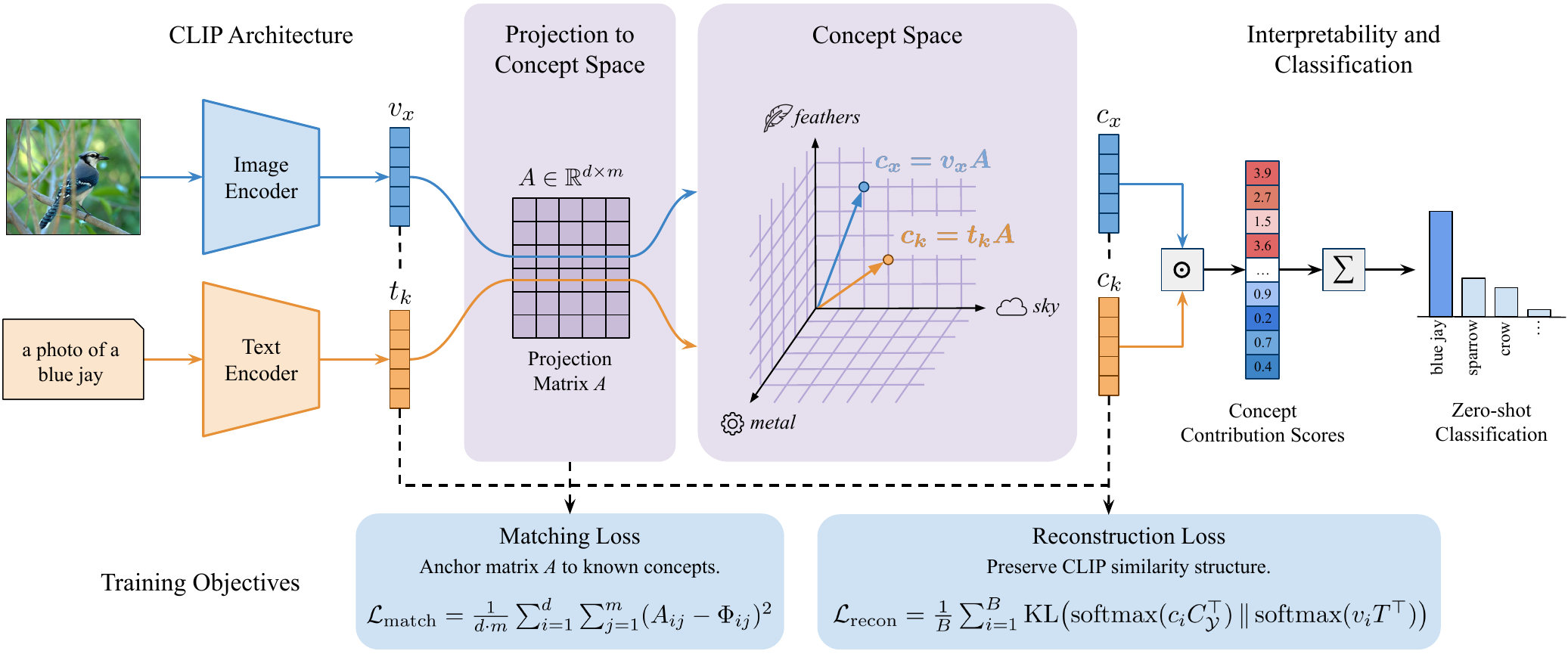}
    \caption{\textbf{Overview of EZPC.} CLIP image and text embeddings are projected into a shared concept space using a learnable matrix $A$. The projected representations $c_x$ and $c_k$ provide (i) concept-based explanations via a Hadamard product and (ii) class logits via a dot-product in concept space. Training jointly optimizes a matching loss and a reconstruction loss to preserve CLIP’s predictive behavior.}
    \label{fig:method_figure}
\end{figure*}

The rapid integration of machine learning into real-world systems has intensified the demand for models that are not only accurate but also transparent and trustworthy. In high-stakes domains such as medical imaging, autonomous driving, and scientific discovery, understanding why a model makes a particular prediction is as critical as the prediction itself. Despite their impressive capabilities, modern deep networks remain largely black boxes, making it difficult to interpret their internal reasoning or diagnose their failures.

Concept Bottleneck Models (CBMs) \citep{pmlr-v119-koh20a} address this issue by introducing an intermediate layer composed of human-understandable concepts. These models decompose the prediction process into two stages: (1) mapping inputs to concept activations, and (2) predicting the final class label based on these activations. This structure provides an interpretable interface between perception and decision-making, allowing users to inspect, validate, or even modify concept activations to understand or correct model behavior.

However, classical CBMs are constrained by two major limitations. First, they require dense concept supervision, which is often expensive or infeasible to collect. Second, they operate under a closed-world assumption: CBMs are trained and evaluated on a fixed set of classes and thus fail to generalize to unseen categories or new domains. These constraints limit their scalability and applicability to open-vocabulary recognition problems. Recent efforts to mitigate these issues by leveraging vision-language models \citep{oikarinenlabel, yang2023language, yuksekgonul2023posthoc} reduce annotation costs, but still require task-specific training and cannot generalize to unseen classes.

In contrast, vision-language models (VLMs) such as CLIP \citep{pmlr-v139-radford21a}, ALIGN \citep{pmlr-v139-jia21b}, and SigLIP \citep{siglip} demonstrate strong open-vocabulary generalization by aligning images and text in a shared semantic space. CLIP, for instance, learns to associate visual and textual information at scale, enabling zero-shot classification by comparing an image embedding with textual embeddings of candidate labels. Without task-specific training, CLIP can accurately recognize objects from natural-language descriptions, which is a significant leap toward flexible, general-purpose perception.

Yet, this generalization comes at the cost of interpretability. CLIP’s embeddings are high-dimensional and entangled, offering little insight into what visual or semantic properties drive a particular decision. As a result, users cannot easily understand why CLIP associates an image with a given label, nor can they trace these predictions back to human-interpretable reasoning.

In this paper, we aim to bridge the gap between the interpretability of CBMs and the generalization ability of CLIP. We propose a method that explains CLIP’s zero-shot predictions through human-understandable concepts. Instead of retraining CLIP or relying on annotated concepts, our method introduces a lightweight decomposition that projects CLIP’s image-text embeddings into a shared concept space. This enables faithful, concept-level explanations of CLIP’s predictions while maintaining its zero-shot capabilities.

Our method, ``Explaining CLIP Zero-shot Predictions Through Concepts" (\textbf{\methodname{}}), aligns CLIP’s representations with a predefined concept basis using two complementary objectives: (i) a matching loss that enforces alignment between learned and known concept embeddings, and (ii) a reconstruction loss that preserves CLIP’s similarity structure in the concept space. The resulting model not only interprets CLIP’s predictions through meaningful concepts but also retains high zero-shot accuracy across diverse datasets.

\myparagraph{Contributions.}
Our key contributions are as follows:
\begin{itemize}
    \item We propose a novel method that decomposes CLIP’s image-text embeddings into a shared concept space, enabling interpretable zero-shot predictions.
    \item We introduce two training objectives, matching and reconstruction, that jointly align concept projections with CLIP’s latent space while preserving semantic fidelity.
    \item We demonstrate through quantitative and qualitative experiments on five benchmarks that \methodname{} provides human-interpretable explanations of CLIP predictions with minimal performance loss.
\end{itemize}

\section{Related Work}
Our work lies at the intersection of zero-shot learning, vision-language modeling, and concept-based interpretability. Below, we review the most relevant research in each area and discuss how our approach relates to prior work.

\myparagraph{Zero-shot Learning.}
Zero-shot learning (ZSL) aims to recognize unseen categories without explicit training data for those classes. Early work achieved this by leveraging human-defined attributes as semantic intermediaries \citep{berkan_demirel,6571196,DAP}, enabling recognition of novel classes through shared attributes. Later approaches replaced attributes with distributed word embeddings such as Word2Vec, GloVe, and BERT to establish semantic correspondences between visual and linguistic spaces \citep{word2vec, pennington2014glove, devlin2019bert}. 

Embedding-based ZSL methods \citep{conse, devise, ale, eszsl} projected both images and labels into a shared latent space, where similarity was computed for classification. Other approaches \citep{Christensen_2023_ICCV, Wang_2018_CVPR, Kampffmeyer_2019_CVPR, Li_2019_ICCV} instead aim to learn weights for zero-shot classification. More recently, large-scale vision-language models such as CLIP \citep{pmlr-v139-radford21a}, ALIGN \citep{pmlr-v139-jia21b}, and SigLIP \citep{siglip} have shown remarkable zero-shot generalization by aligning visual and textual modalities through contrastive learning on massive image-text datasets. These models removed the need for explicit semantic attribute design and established a new paradigm for open-vocabulary recognition.

\myparagraph{Vision-Language Models.}
Vision-language models jointly learn image and text representations in a shared embedding space. CLIP’s contrastive learning objective allows image embeddings to align with corresponding text embeddings, enabling zero-shot recognition of arbitrary categories described in natural language. Subsequent works such as CoOp \citep{zhou2022cocoop}, BLIP \citep{li2022blip}, and PaLI-Gemma \citep{beyer2024paligemma} expanded the scale and adaptability of VLMs, incorporating prompt tuning, caption generation, and cross-modal retrieval.

Beyond classification, VLMs have become a powerful foundation for interpretability studies. Their multimodal embeddings naturally encode high-level semantics that can be probed with linguistic queries or concept prompts \citep{oikarinenlabel, Rao2024Discover}. However, these models lack explicit interpretability mechanisms; while they capture semantics implicitly, the internal representations remain opaque and difficult to translate into human-understandable concepts.

\myparagraph{Concept Bottleneck Models.}
Concept Bottleneck Models \citep{pmlr-v119-koh20a} improve interpretability by decomposing predictions into human-defined concepts. A CBM first predicts interpretable attributes (e.g., \emph{has wings, is red}) and then uses these attributes for downstream classification. This design provides transparency and controllability: users can inspect or modify the concept activations to understand and alter model decisions. 

Subsequent works have enhanced CBMs by improving their robustness \citep{yan2023robust, kim2023concept, yuksekgonul2023posthoc, incr_residual_cbm}, discovering latent concepts automatically \citep{Rao2024Discover}, or integrating textual guidance from large language models \citep{yu2025language, oikarinenlabel, srivastava2024vlg, yang2023language}. Despite these advances, most CBMs remain limited to closed-world settings, requiring concept supervision and predefined label sets.

\myparagraph{CLIP Interpretability and Zero-shot CBMs.}
Recent works \citep{menon2022visual, Yan_2023_ICCV} have sought to merge CLIP’s open-vocabulary recognition with the interpretability of concept-based reasoning. Gandelsman et al. \citep{gandelsman2024interpreting} decomposes CLIP’s image encoder across patches, layers, and attention heads, using CLIP’s text encoder to interpret component-wise contributions. SpLiCE \citep{NEURIPS2024_996bef37} decomposes CLIP’s embeddings into sparse linear combinations of concept vectors, producing per-instance concept explanations but at the cost of expensive test-time optimization for each image. Z-CBM \citep{yamaguchi2025zero} extends the CBM paradigm to the zero-shot setting by reconstructing embeddings from concept banks, yet it depends on large concept repositories and costly regressions.

In contrast, \methodname{} introduces a unified, trainable decomposition that aligns CLIP’s vision-language space with a shared, interpretable concept basis. Unlike prior instance-specific or retrieval-based methods, \methodname{} learns a single projection that jointly preserves semantic alignment and interpretability, yielding efficient, faithful explanations of zero-shot predictions.

\section{Method}
We propose a method that explains CLIP’s zero-shot predictions through a human-interpretable concept space. Our method learns a linear decomposition of CLIP’s joint image-text embeddings into concept activations. This decomposition preserves the semantic relationships inherent in CLIP’s representation while making them interpretable.

\subsection{Concept Decomposition} \label{subsec:concept_decomposition}
Let $\mathcal{Y} = \{y_1, \dots, y_K\}$ denote a set of $K$ candidate class labels. We define a learnable projection matrix $A \in \mathbb{R}^{d \times m}$ that should map CLIP's $d$-dimensional embedding space into a concept space with $m$ interpretable dimensions. For this, we want each of the $m$ columns of $A$ to correspond to a distinct, human-interpretable concept direction in CLIP's space (e.g., \emph{feathers, metal, sky}). We describe how this is achieved through initialization and training in Section~\ref{subsec:training_objectives}.

Now, let $f_{\text{img}}$ and $f_{\text{text}}$ denote CLIP's image and text encoders, respectively, and define the normalized embeddings
\begin{equation}
    v_x = \frac{f_{\text{img}}(x)}{\|f_{\text{img}}(x)\|}, \qquad
    t_k = \frac{f_{\text{text}}(y_k)}{\|f_{\text{text}}(y_k)\|},
\end{equation}
where $v_x \in \mathbb{R}^d$ is the image embedding and $t_k \in \mathbb{R}^d$ is the text embedding of the $k$-th class label. We stack the class embeddings into a matrix $T = [t_1; \dots; t_K] \in \mathbb{R}^{K \times d}$.

Using $A$, we then compute concept activations for images and labels in the shared concept space as
\begin{equation}
    c_x = v_x A, \qquad C_\mathcal{Y} = T A,
\end{equation}
where $c_x \in \mathbb{R}^m$ and $C_\mathcal{Y} \in \mathbb{R}^{K \times m}$.
Each row of $C_\mathcal{Y}$ gives the concept activations of the corresponding class label.

\subsection{Training Objectives} \label{subsec:training_objectives}
Our goal is to learn a concept projection that preserves CLIP's similarity structure while remaining interpretable. For this, we jointly optimize two complementary objectives.

\myparagraph{(1) Matching Loss.}
We initialize $A$ from a set of concept embeddings relevant to the target domain, such as visual attributes. 
For a set of $m$ concepts, we compute and stack their CLIP text embeddings to form a matrix $\Phi \in \mathbb{R}^{d \times m}$, such that each column corresponds to a concept phrase (e.g., \emph{has feathers, made of metal}).
We initialize $A = \Phi$ and use a mean-squared matching loss to keep $A$ close to this interpretable basis throughout training:
\begin{equation}
    \mathcal{L}_{\text{match}} = \frac{1}{d \cdot m} \sum_{i=1}^{d}\sum_{j=1}^{m}
    (A_{ij} - \Phi_{ij})^2.
\end{equation}
This anchoring ensures that the columns of $A$ remain aligned with known concept directions, preserving interpretability.

\myparagraph{(2) Reconstruction Loss.}
To ensure that the decomposition preserves CLIP's zero-shot similarity structure, we introduce a reconstruction loss based on the KL divergence between the original CLIP similarity distribution and the concept-based distribution. Given a batch of $B$ image embeddings $\{v_i\}_{i=1}^B$, we define:
\begin{equation}
    \mathcal{L}_{\text{recon}} =
    \frac{1}{B}\sum_{i=1}^{B}
    \text{KL}\!\left(
        \text{softmax}(c_i \, C_\mathcal{Y}^\top)
        \,\|\,
        \text{softmax}(v_i T^\top)
    \right),
\end{equation}
where $c_i = v_i A$ and $C_\mathcal{Y} = TA$ are the concept activations defined in Section~\ref{subsec:concept_decomposition}. This enforces that the concept-space similarity distribution remains consistent with CLIP's original predictions, ensuring semantic faithfulness.

\myparagraph{(3) Total Loss.}
The overall objective combines both terms with a balancing coefficient $\lambda$:
\begin{equation}
    \mathcal{L}_{\text{total}} = \mathcal{L}_{\text{match}} + \lambda \, \mathcal{L}_{\text{recon}}.
\end{equation}

\subsection{Concept-based Inference}
At inference, we perform zero-shot classification directly in the concept space:
\begin{equation}
    \hat{y} = \underset{k \in \lbrace 1,...,K\rbrace}{\arg\max} \, \langle c_x, c_k \rangle,
\end{equation}
where $c_k$ denotes the $k$-th row of $C_\mathcal{Y}$. We define a vector of concept-wise interaction scores between image $x$ and class $y_k$ as
\begin{equation}
    s_{x,k} := c_x \odot c_k,
\end{equation}
where the $j$-th entry of $s_{x,k}$ is large when both the image and the class strongly activate the $j$-th concept. The overall concept-space similarity decomposes as
\begin{equation}
    \langle c_x, c_k \rangle = \sum_{j=1}^m s_{x,k}^{(j)}.
\end{equation}
Each dimension of $s_{x,k}$ quantifies how strongly a specific concept contributes to the image-text alignment. Since the concept scores compose the prediction logit directly, the explanations provided by \methodname{} are faithful to the model's decision process by construction. Thus, the model not only classifies unseen classes in a zero-shot manner but also provides fine-grained, faithful explanations identifying which concepts drive its decisions.

\section{Experiments}
\label{sec:experiments}

\myparagraph{Datasets.}
We evaluate our approach on five benchmark datasets covering diverse visual domains and levels of granularity: CIFAR-100~\cite{Krizhevsky09learningmultiple}, CUB-200-2011 (CUB)~\cite{wah2011caltech}, ImageNet-100~\cite{imagenet15russakovsky}, ImageNet-1k~\cite{imagenet15russakovsky}, and Places365~\cite{zhou2017places}.  
Each dataset is partitioned into seen and unseen classes following an $80/20$ split, ensuring that unseen categories are completely excluded during training. This setup enables evaluation under both the zero-shot (unseen-only) and generalized zero-shot (joint seen-unseen) settings.

The chosen datasets span a wide range of visual abstraction levels: CIFAR-100 and ImageNet-100 capture compact, object-centric imagery; CUB offers fine-grained bird species classification; ImageNet-1k provides large-scale diversity; and Places365 focuses on scene-level understanding. This diversity allows us to assess the interpretability and generalization of \methodname{} across varying domains.

\myparagraph{Concept space.}  
We adopt the human-interpretable concept sets introduced by LF-CBM~\cite{oikarinenlabel}, where concepts were originally generated using GPT-3~\cite{gpt3} based on the class names in each dataset. Specifically, the concept vocabulary sizes are as follows: CIFAR-100 (\emph{892 concepts}), CUB (\emph{370 concepts}), ImageNet-1k (\emph{4,751 concepts}), and Places365 (\emph{2,544 concepts}). Since CIFAR-100, CUB, and Places365 contain relatively few concepts, we merge their original concept vocabularies with ImageNet-1k’s larger concept pool to obtain a richer and more transferable set of interpretable attributes. This unified concept space ensures consistent coverage of visual semantics across datasets and provides a fair basis for cross-domain comparison in our zero-shot evaluation.

\begin{table*}[t]
    \centering
    \caption{\textbf{Generalized zero-shot performance across five datasets.} Each dataset is partitioned into seen (80\%) and unseen (20\%) classes. All models use the CLIP RN50 backbone. \methodname{} retains strong performance compared to CLIP while introducing interpretability.}
    \label{tab:multi-dataset-results}
    \resizebox{\textwidth}{!}{%
    \renewcommand{\arraystretch}{1.2}
    \begin{tabular}{l|
                    S[table-format=1.3] S[table-format=1.3] S[table-format=1.3]
                    S[table-format=1.3] S[table-format=1.3] S[table-format=1.3]
                    S[table-format=1.3] S[table-format=1.3] S[table-format=1.3]
                    S[table-format=1.3] S[table-format=1.3] S[table-format=1.3]
                    S[table-format=1.3] S[table-format=1.3] S[table-format=1.3]} 
        \toprule
         & \multicolumn{3}{c}{\textbf{CIFAR-100}} 
         & \multicolumn{3}{c}{\textbf{ImageNet-100}} 
         & \multicolumn{3}{c}{\textbf{CUB}} 
         & \multicolumn{3}{c}{\textbf{ImageNet-1k}} 
         & \multicolumn{3}{c}{\textbf{Places365}} \\
        \cmidrule(lr){2-4} \cmidrule(lr){5-7} \cmidrule(lr){8-10} \cmidrule(lr){11-13} \cmidrule(lr){14-16}
        \textbf{Model} & {Seen} & {Unseen} & {H} 
                       & {Seen} & {Unseen} & {H} 
                       & {Seen} & {Unseen} & {H} 
                       & {Seen} & {Unseen} & {H} 
                       & {Seen} & {Unseen} & {H} \\
        \midrule
        \textbf{CLIP~\cite{pmlr-v139-radford21a}}          & 0.370 & 0.454 & 0.408 & 0.680 & 0.707 & 0.693 & 0.468 & 0.481 & 0.474 & 0.513 & 0.548 & 0.530 & 0.350 & 0.375 & 0.362 \\ \hline
        \textbf{Z-CBM~\cite{yamaguchi2025zero}}         & 0.319 & 0.425 & 0.365 & 0.592 & 0.579 & 0.585 & 0.183 & 0.195 & 0.189 & 0.439 & 0.486 & 0.462 & \textbf{0.349} & 0.365 & \textbf{0.357} \\
        \textbf{SpLiCE~\cite{NEURIPS2024_996bef37}} & 0.248 & 0.298 & 0.270 & 0.371 & 0.409 & 0.389 & 0.100 & 0.053 & 0.070 & 0.275 & 0.331 & 0.300 & 0.276 & 0.288 & 0.282 \\
        \rowcolor{blue!8}
        \textbf{\methodname{}} & \textbf{0.365} & \textbf{0.449} & \textbf{0.403} & \textbf{0.675} & \textbf{0.690} & \textbf{0.682} & \textbf{0.457} & \textbf{0.473} & \textbf{0.465} & \textbf{0.468} & \textbf{0.494} & \textbf{0.481} & 0.339 & \textbf{0.366} & 0.352 \\
        \bottomrule
    \end{tabular}%
    }
\end{table*}

\myparagraph{Evaluation Metrics.}
In the standard zero-shot setting, given an image $x$ and a set of class names $\mathcal{Y} = \{y_1, y_2, \dots, y_K\}$, the prediction is made by selecting the class whose text embedding has the highest cosine similarity with the image embedding
\begin{equation}
\hat{y} = \underset{y_k \in \mathcal{Y}}{\arg\max} \langle v_x, t_k \rangle,
\end{equation}
where $v_x$ and $t_k$ denote the normalized CLIP embeddings for the image and text, respectively.  

In the generalized zero-shot (GZS) setting, the label space includes both seen and unseen classes,  
$\mathcal{Y}_{\text{GZS}} = \mathcal{Y}_{\text{seen}} \cup \mathcal{Y}_{\text{unseen}}$.  
We report accuracies on seen ($\mathrm{Acc}_S$) and unseen ($\mathrm{Acc}_U$) classes, and use their harmonic mean ($H$) as a balanced measure which penalizes models that overfit to either seen or unseen categories and thus reflects overall generalization capability.

\myparagraph{Baselines.}
We compare \methodname{} against three baselines:
(1) the original CLIP~\cite{pmlr-v139-radford21a},  
(2) the zero-shot concept bottleneck variant Z-CBM~\cite{yamaguchi2025zero}, and  
(3) the sparse linear decomposition approach SpLiCE~\cite{NEURIPS2024_996bef37}.  
All models are evaluated with identical backbones to ensure comparability using their official implementations with default hyperparameters.

For all experiments, except the backbone ablation, we use the CLIP RN50 backbone, as it is widely adopted in prior work. Unless otherwise noted, ablations are conducted on ImageNet-100, which provides a convenient testbed for evaluating different settings. We use $\lambda=1$ for all datasets, except CUB and Places365, where $\lambda=5$ gives better quantitative and qualitative results.

\subsection{Quantitative Analysis}

\myparagraph{Results.} 
Table~\ref{tab:multi-dataset-results} reports generalized zero-shot performance across the five datasets. On CIFAR-100, ImageNet-100, and CUB, \methodname{} remains within roughly 1\% harmonic mean of CLIP, showing that most of CLIP’s accuracy is retained despite the addition of an interpretable concept layer. On ImageNet-1k, the gap is larger (around 5\%), reflecting the increased difficulty of this large-scale setting. For Places365, \methodname{} is about 1\% below CLIP and close to Z-CBM. Across the remaining datasets, \methodname{} provides clear improvements over Z-CBM and SpLiCE, often exceeding them by 10-15\% in harmonic mean. Overall, these results demonstrate that \methodname{} offers meaningful interpretability while keeping performance competitive with CLIP and substantially stronger than prior concept-based baselines.

\myparagraph{Backbone sensitivity.}
As summarized in Table~\ref{tab:backbone-ablation}, larger and more expressive backbones improve performance for \methodname{}. This indicates that the concept-based decomposition scales naturally with model capacity, maintaining interpretability across different architectures.

\begin{table}
    \centering
    \caption{\textbf{Effect of backbone architecture on zero-shot and generalized zero-shot performance.}
    Larger backbones consistently improve both zero-shot and generalized zero-shot performance.}
    \label{tab:backbone-ablation}
    \renewcommand{\arraystretch}{1.1}
    \setlength{\tabcolsep}{3pt}
    \scriptsize
    \begin{tabular}{llccccc}
        \toprule
        \textbf{Backbone} & \textbf{Variant} & \multicolumn{2}{c}{\textbf{Zero-shot}} & \multicolumn{3}{c}{\textbf{Generalized}} \\
        \cmidrule(lr){3-4} \cmidrule(lr){5-7}
         &  & Seen & Unseen & Seen & Unseen & H \\
        \midrule
        \multirow{2}{*}{\centering CLIP RN50}
          & Base
          & 0.706 & 0.855 & 0.680 & 0.707 & 0.693 \\
          & \cellcolor{blue!8}\methodname{}
          & \cellcolor{blue!8}0.699
          & \cellcolor{blue!8}0.851
          & \cellcolor{blue!8}0.675
          & \cellcolor{blue!8}0.690
          & \cellcolor{blue!8}0.682 \\
        \midrule
        \multirow{2}{*}{\centering CLIP ViT-B/32}
          & Base
          & 0.729 & 0.887 & 0.703 & 0.715 & 0.709 \\
          & \cellcolor{blue!8}\methodname{}
          & \cellcolor{blue!8}0.724
          & \cellcolor{blue!8}0.879
          & \cellcolor{blue!8}0.694
          & \cellcolor{blue!8}0.716
          & \cellcolor{blue!8}0.705 \\
        \midrule
        \multirow{2}{*}{\centering CLIP ViT-L/14}
          & Base
          & 0.839 & 0.925 & 0.821 & 0.836 & 0.828 \\
          & \cellcolor{blue!8}\methodname{}
          & \cellcolor{blue!8}0.832
          & \cellcolor{blue!8}0.924
          & \cellcolor{blue!8}0.812
          & \cellcolor{blue!8}0.831
          & \cellcolor{blue!8}0.822 \\
        \midrule
        \multirow{2}{*}{\centering SigLIP ViT-SO400M/14}
          & Base
          & 0.882 & 0.972 & 0.871 & 0.889 & 0.880 \\
          & \cellcolor{blue!8}\methodname{}
          & \cellcolor{blue!8}0.880
          & \cellcolor{blue!8}0.972
          & \cellcolor{blue!8}0.870
          & \cellcolor{blue!8}0.886
          & \cellcolor{blue!8}0.878 \\
        \bottomrule
    \end{tabular}
\end{table}

\begin{figure}
  \centering

  \begin{subfigure}{0.85\columnwidth}
      \centering
      \includegraphics[width=\columnwidth]{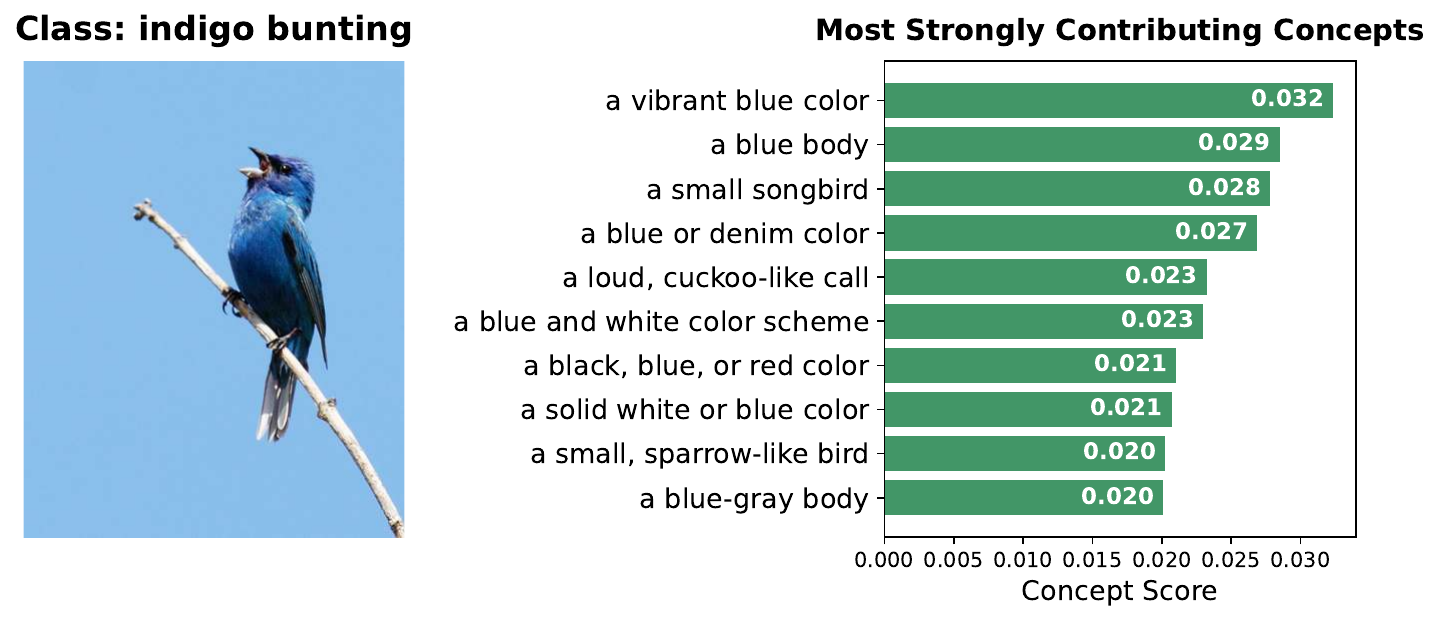}
      \caption{Image-level analysis for $\lambda = 1$.}
  \end{subfigure}

  \begin{subfigure}{0.85\columnwidth}
      \centering
      \includegraphics[width=\columnwidth]{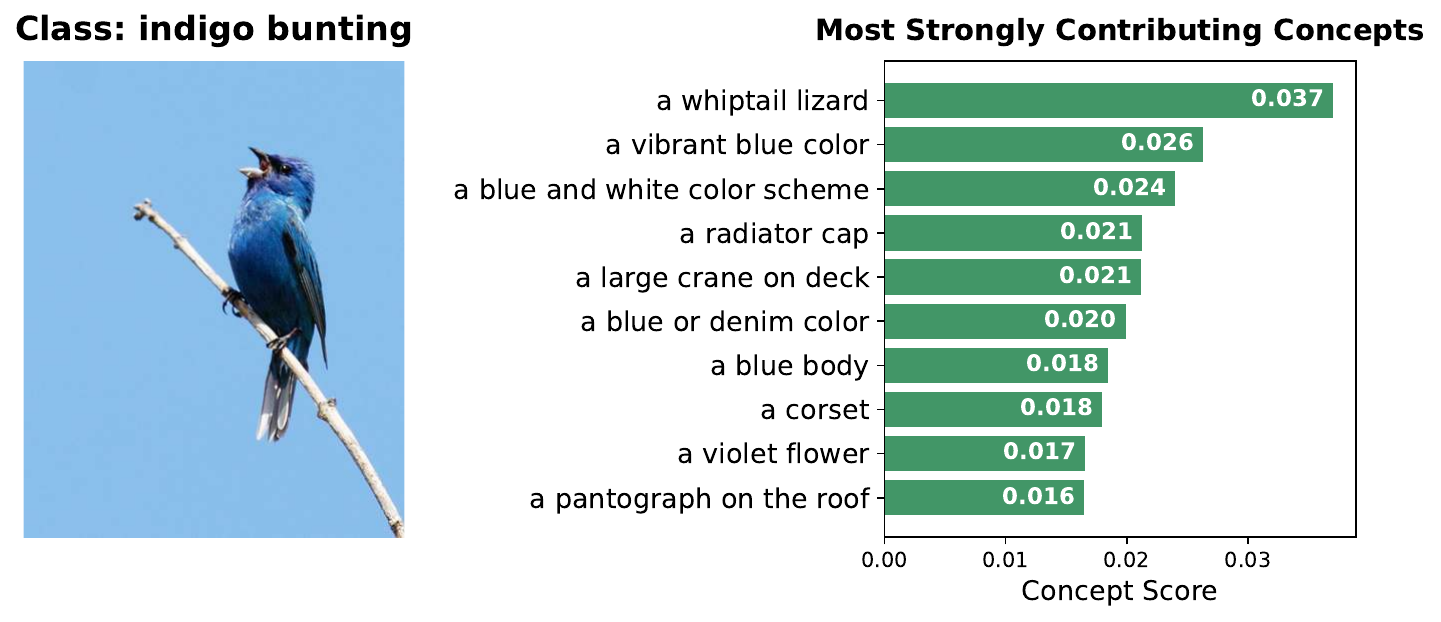}
      \caption{Image-level analysis for $\lambda = 100$.}
  \end{subfigure}

    \caption{\textbf{Qualitative comparison of image-level explanations for different $\lambda$ values.} For $\lambda=1$, \methodname{} produces semantically consistent concept activations. For larger values (e.g., $\lambda=100$), unrelated concepts appear among the top activations.}
  \label{fig:lambda_comparison}
\end{figure}

\myparagraph{Effect of $\lambda$.}
Table~\ref{tab:lambda-ablation} shows that larger $\lambda$ values improve quantitative performance by emphasizing the reconstruction loss, which better preserves CLIP’s similarity structure. However, Figure~\ref{fig:lambda_comparison} reveals the opposite trend qualitatively: $\lambda=1$ produces image-relevant concepts, whereas higher values (e.g., $\lambda=100$) introduce unrelated activations. This trade-off is expected, smaller $\lambda$ strengthens the matching loss, keeping learned concept directions closer to CLIP’s concept embeddings and yielding more interpretable explanations.

\begin{table}
    \tiny
    \centering
    \caption{\textbf{Effect of the reconstruction weighting parameter $\lambda$.} Larger $\lambda$ values improve both zero-shot and generalized performance, with moderate to high settings giving the best results.}
    \label{tab:lambda-ablation}
    \resizebox{\columnwidth}{!}{%
    \begin{tabular}{l
                    S[table-format=1.3]
                    S[table-format=1.3]
                    S[table-format=1.3]
                    S[table-format=1.3]
                    S[table-format=1.3]}
        \toprule
         & \multicolumn{2}{c}{\textbf{Zero-shot}} & \multicolumn{3}{c}{\textbf{Generalized Zero-shot}} \\
        \cmidrule(lr){2-3} \cmidrule(lr){4-6}
        $\boldsymbol{\lambda}$ & {Seen} & {Unseen} & {Seen} & {Unseen} & {H} \\
        \midrule
        \textbf{0.01} & 0.377 & 0.508 & 0.347 & 0.371 & 0.358 \\
        \textbf{0.1} & 0.654 & 0.820 & 0.626 & 0.633 & 0.630 \\
        \textbf{1} & 0.699 & 0.851 & 0.675 & 0.690 & 0.682 \\
        \textbf{10} & 0.707 & 0.859 & 0.681 & 0.709 & 0.695 \\
        \textbf{100} & 0.706 & 0.857 & 0.680 & 0.704 & 0.692 \\
        \textbf{1000} & 0.707 & 0.857 & 0.680 & 0.708 & 0.694 \\
        \bottomrule
    \end{tabular}%
    }
\end{table}

\subsection{Cross-Dataset Experiments}
To rigorously assess the generalization ability of our model beyond its training distribution, we perform cross-dataset experiments where the concept bottleneck is trained on a source dataset (ImageNet-100) and evaluated on distinct target datasets (CIFAR-100 and CUB) without any fine-tuning. This setup examines whether the learned projection matrix $A$ captures transferable semantics that remain valid across domains with different visual granularity and label taxonomies.

\myparagraph{Setup.}
During evaluation, we treat classes from the source dataset (e.g., ImageNet-100) as \emph{seen} and classes from the target dataset (e.g., CUB) as \emph{unseen}. For the generalized zero-shot setting, we merge all categories from both datasets and jointly predict across this combined label space, a substantially more challenging scenario than standard zero-shot transfer. We report \emph{Seen} and \emph{Unseen} accuracy under both standard zero-shot and generalized zero-shot settings.

\myparagraph{Results.}
Table~\ref{tab:cross-in100-merged} reports cross-dataset transfer performance when the projection is trained on ImageNet-100 and evaluated on CIFAR-100 and CUB.  
For CIFAR-100, \methodname{} produces zero-shot and generalized zero-shot accuracies that are close to CLIP: the seen accuracies differ by less than 0.5\%, and the unseen accuracies are within roughly 2-3\%. In the generalized setting, \methodname{} achieves a harmonic mean about 3\% higher than CLIP. For CUB, the differences are similarly small: \methodname{} is within about 1-2\% of CLIP on both seen and unseen zero-shot accuracies, and the harmonic mean differs by roughly 1\%. These results indicate that the concept projection learned from ImageNet-100 transfers reasonably well to both object-centric and fine-grained domains, maintaining performance close to CLIP without any fine-tuning.

\begin{table}
    \centering
    \caption{\textbf{Cross-dataset transfer results for \methodname{} trained on ImageNet-100 compared to CLIP.} We evaluate on two target datasets: CIFAR-100 and CUB. \emph{Seen} classes correspond to ImageNet-100 categories, while \emph{unseen} classes correspond to the target dataset.}
    \label{tab:cross-in100-merged}
    \resizebox{\columnwidth}{!}{%
    \begin{tabular}{l
                    l
                    S[table-format=1.3]
                    S[table-format=1.3]
                    S[table-format=1.3]
                    S[table-format=1.3]
                    S[table-format=1.3]}
        \toprule
        \textbf{Target Dataset} & \textbf{Model} 
        & \multicolumn{2}{c}{\textbf{Zero-shot}} 
        & \multicolumn{3}{c}{\textbf{Generalized Zero-shot}} \\
        \cmidrule(lr){3-4} \cmidrule(lr){5-7}
         &  & {Seen} & {Unseen} & {Seen} & {Unseen} & {H} \\
        \midrule

        \multirow{2}{*}{\textbf{CIFAR-100}}
            & CLIP            & 0.686 & 0.387 & 0.663 & 0.266 & 0.380 \\
            & \cellcolor{blue!8}\methodname{}
                                & \cellcolor{blue!8}0.684 
                                & \cellcolor{blue!8}0.363 
                                & \cellcolor{blue!8}0.659 
                                & \cellcolor{blue!8}0.296 
                                & \cellcolor{blue!8}0.409 \\
        \midrule

        \multirow{2}{*}{\textbf{CUB}}
            & CLIP            & 0.686 & 0.471 & 0.617 & 0.458 & 0.526 \\
            & \cellcolor{blue!8}\methodname{}
                                & \cellcolor{blue!8}0.674 
                                & \cellcolor{blue!8}0.461 
                                & \cellcolor{blue!8}0.607 
                                & \cellcolor{blue!8}0.448 
                                & \cellcolor{blue!8}0.515 \\
        \bottomrule
    \end{tabular}%
    }
\end{table}

\subsection{Qualitative Analysis}

\subsubsection{Image-level Explanations}
\begin{figure*}
  \centering
  \begin{subfigure}{0.89\textwidth}
      \centering
      \includegraphics[width=\textwidth]{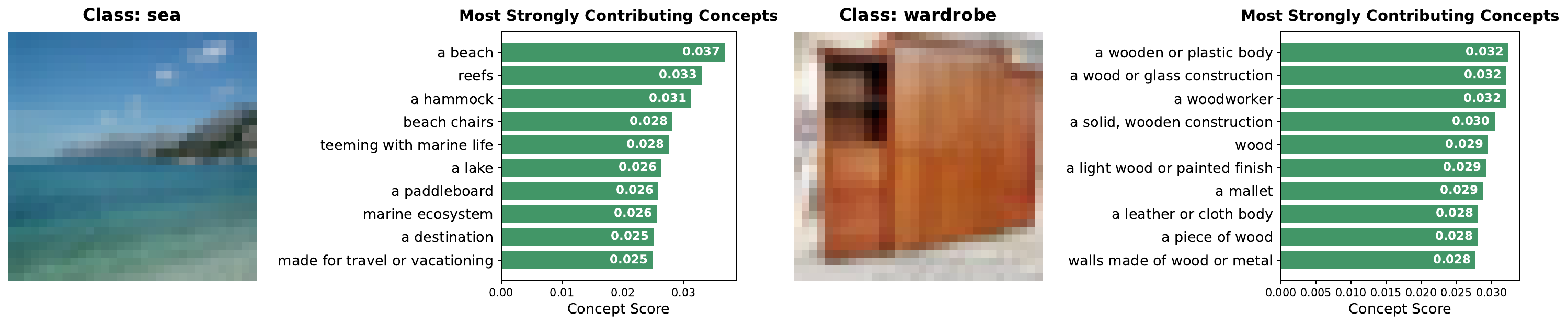}
      \caption{CIFAR-100 image-level explanations.}
      \label{fig:image-level-cifar100}
  \end{subfigure}
  \begin{subfigure}{0.89\textwidth}
      \centering
      \includegraphics[width=\textwidth]{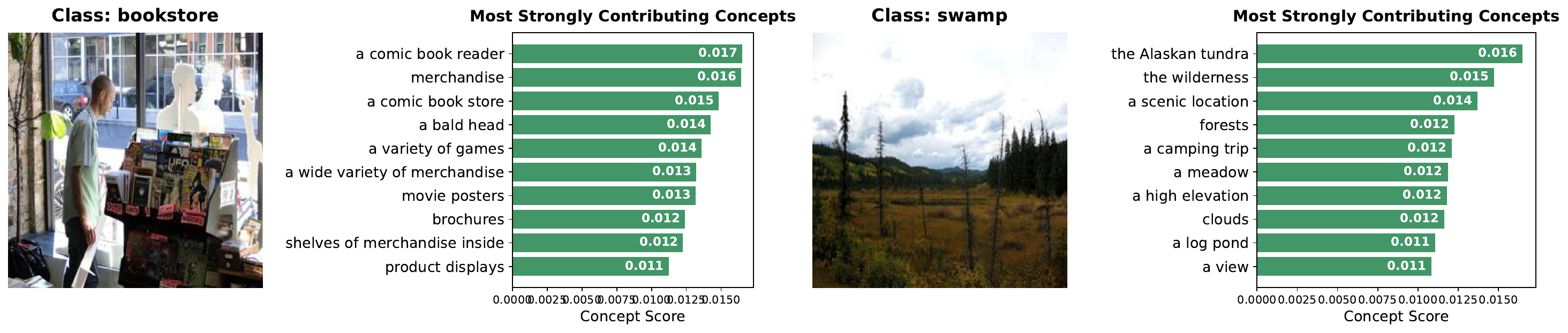}
      \caption{Places365 image-level explanations.}
      \label{fig:image-level-places365}
  \end{subfigure}
    \caption{\textbf{Image-level Explanations.} For each image, EZPC displays the top-10 activated concepts that contribute most to the zero-shot prediction. The highlighted concepts closely correspond to salient visual characteristics of the input images.
    }
  \label{fig:image-level-all}
\end{figure*}

For image-level explanations, EZPC follows a straightforward procedure: it projects both the image embedding and the predicted class label embedding into the concept space, computes their element-wise product, and selects the top-10 activated concepts that drive the zero-shot prediction. Figure~\ref{fig:image-level-cifar100} shows examples from CIFAR-100: for the class \emph{sea}, top concepts include \emph{a beach}, \emph{reefs}, and \emph{a hammock}, while \emph{wardrobe} activates \emph{a wooden or plastic body} and \emph{wood}. Figure~\ref{fig:image-level-places365} shows Places365 examples, where scene-level concepts such as \emph{the Alaskan tundra} and \emph{the wilderness} emerge for the class \emph{swamp}. While most activated concepts are semantically relevant, some reflect class-level associations rather than visual content of the specific image: for \emph{sea}, concepts like \emph{a hammock} and \emph{a paddleboard} are related to the class but not visible in the image.

\subsubsection{Class-level Explanations}
\begin{figure*}
  \centering
  \resizebox{0.88\textwidth}{!}{%
    \begin{minipage}{\textwidth}
      \begin{subfigure}{0.48\textwidth}
          \centering
          \includegraphics[width=\textwidth]{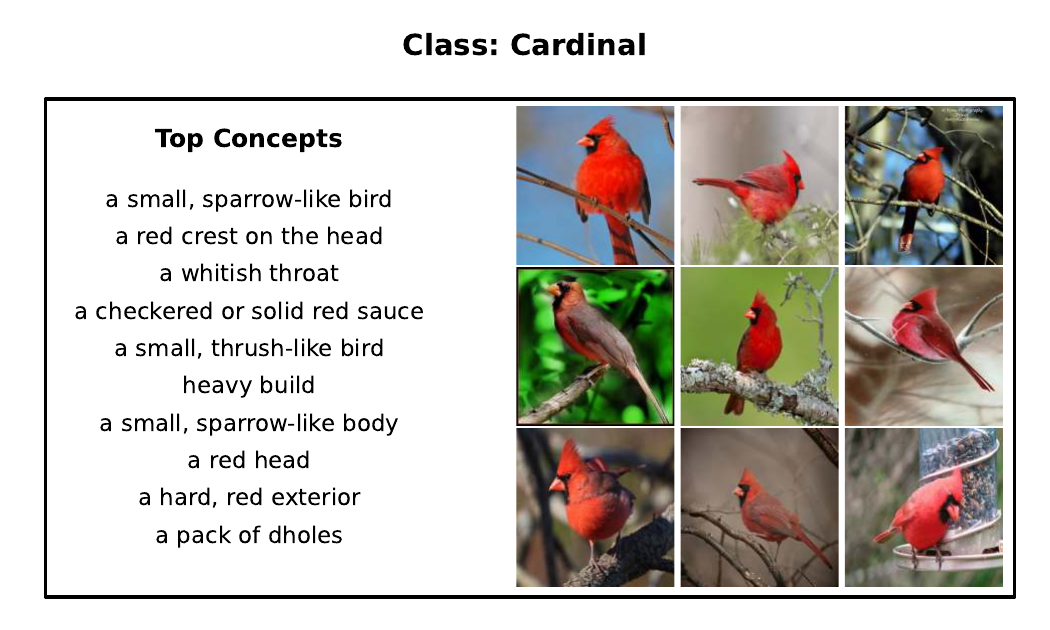}
          \caption{CUB class-level concept explanations.}
      \end{subfigure}
      \hfill
      \begin{subfigure}{0.48\textwidth}
          \centering
          \includegraphics[width=\textwidth]{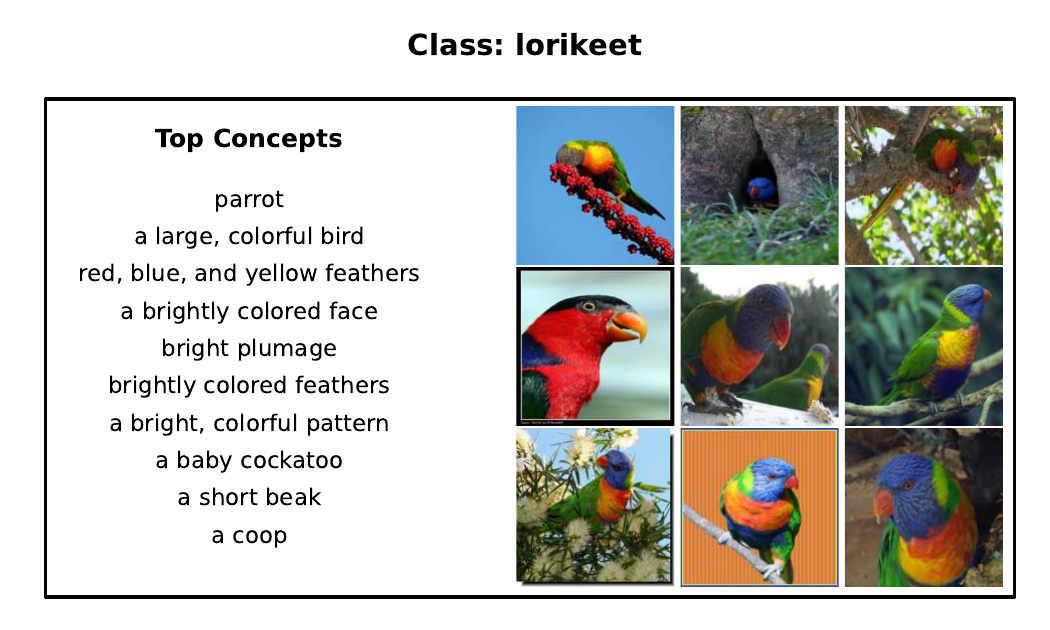}
          \caption{ImageNet-100 class-level concept explanations.}
      \end{subfigure}
    \end{minipage}
  }

    \caption{
    \textbf{Class-level Concept Explanations.}
    For each class, we average concept activations over nine sampled images.  
    \methodname{} produces coherent class signatures, highlighting concepts that characterize each category.
    }
  \label{fig:class_level_explanations_grid}
\end{figure*}

To obtain class-level explanations, we randomly sample nine images from a target class and compute the mean concept activations across these images. This averaged concept profile highlights the concepts that are most representative of the class as a whole, rather than a single instance. Figure~\ref{fig:class_level_explanations_grid} presents such examples from CUB and ImageNet-100. For instance, the CUB class \emph{Cardinal} activates concepts such as \emph{a red crest on the head} and \emph{a red head}, while the ImageNet-100 class \emph{Lorikeet} highlights \emph{parrot}, \emph{red, blue, and yellow feathers}, and \emph{brightly colored feathers}. However, not all activated concepts are meaningful: for Cardinal, concepts such as \emph{heavy build}, \emph{a hard, red exterior}, and \emph{a pack of dholes} appear irrelevant. Such cases may stem from noise in CLIP's embedding space, limitations of the concept vocabulary, etc.

\subsubsection{Concept Clustering}

\begin{figure*}
  \centering
  \begin{subfigure}{0.48\textwidth}
      \centering
      \includegraphics[width=\textwidth]{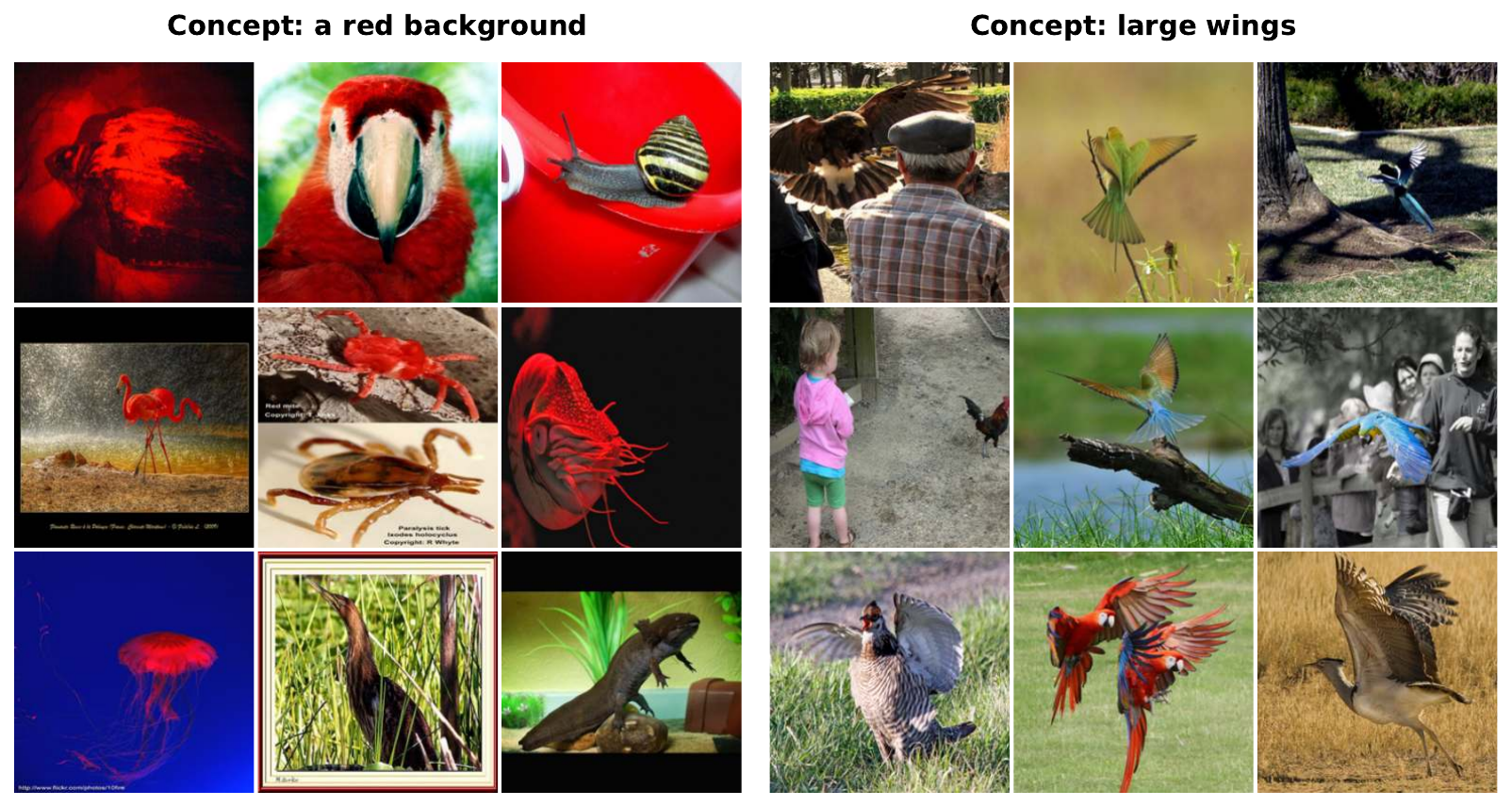}
      \caption{ImageNet-100 concept clustering results.}
    \end{subfigure}
    \hfill
    \begin{subfigure}{0.48\textwidth}
        \centering
          \includegraphics[width=\textwidth]{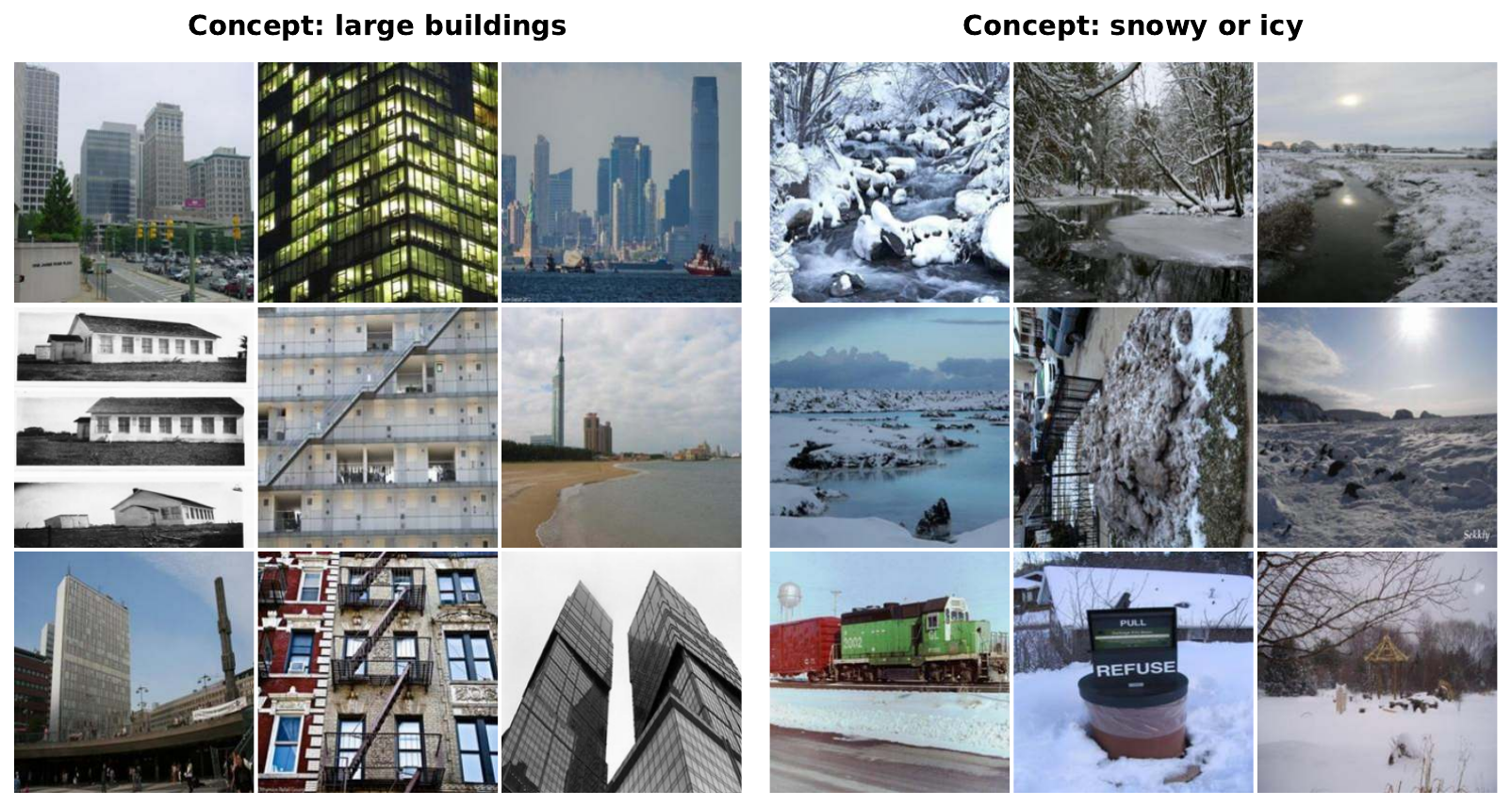}
          \caption{Places365 concept clustering results.}
    \end{subfigure}
    \caption{\textbf{Concept Clustering Results.} For each concept, we retrieve the nine images with the highest activation. Clusters from ImageNet-100 and Places365 show coherent semantic structure, indicating that \methodname{} learns interpretable concept directions.}
\label{fig:concept_clustering_results}
\end{figure*}

We further analyze the semantic structure of the learned concept space by performing concept-based image retrieval. For a given concept, we compute the image-side concept activation for every image in the dataset and retrieve the nine images with the highest activation. This procedure reveals whether individual concept dimensions correspond to visually coherent directions in the embedding space. Figure~\ref{fig:concept_clustering_results} shows example clusters from ImageNet-100 and Places365. For the concept \emph{a red background}, the retrieved images contain prominent red tones; \emph{large wings} retrieves bird images; and \emph{large buildings} retrieves architectural scenes. The strong thematic consistency across retrieved images confirms that individual concept dimensions capture interpretable and visually grounded semantics, rather than encoding arbitrary or entangled directions in CLIP's latent space.

\subsubsection{Concept-Region Alignment}
\begin{figure}[H]
    \centering
    \includegraphics[width=\linewidth]
        {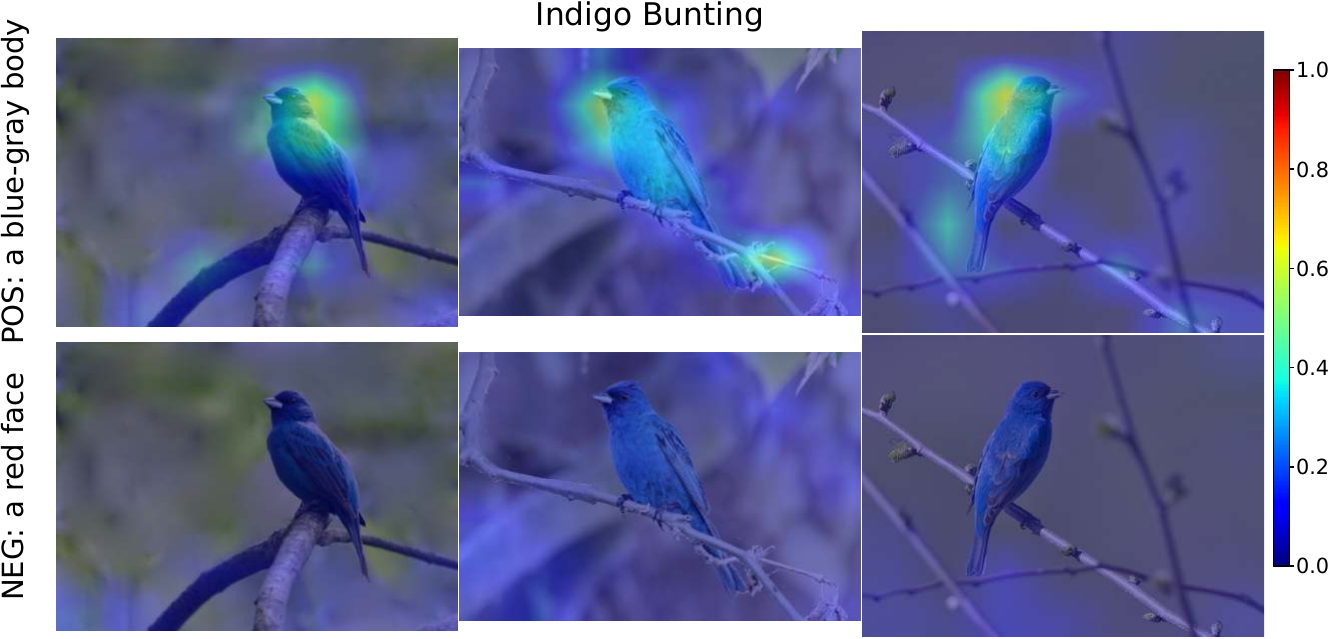}
        \caption{\textbf{Region-level concept alignment for Indigo Bunting.} Top row: positive concept (\emph{a blue-gray body}). Bottom row: negative concept (\emph{a red face}).}
    \label{fig:indigo_bunting_region_alignment}
\end{figure}

To evaluate whether the learned concept space produces spatially meaningful explanations, we analyze region-level alignment between concept activations and object locations. We extract patch-level features from the CLIP RN50 backbone, project them into the concept space via $A$, and visualize the resulting spatial activation maps. Figure~\ref{fig:indigo_bunting_region_alignment} shows an example on CUB, where a positive concept (\emph{a blue-gray body}) produces high activations localized on the bird, while a negative concept (\emph{a red face}) shows near-zero activation.

\vspace{-10pt}

\vspace{-10pt}

We further quantify localization quality on the Indigo Bunting class from CUB using ground-truth segmentation masks. We evaluate a positive concept (\emph{a blue-gray body}) and a negative concept (\emph{a red face}) across all images of this class, computing \emph{Pointing Accuracy}, \emph{Inside Activation Ratio}, \emph{IoU@10\%}, and \emph{IoU@20\%}. As shown in Table~\ref{tab:concept_region_alignment_table}, the positive concept achieves 96.7\% pointing accuracy and substantially higher IoU scores than the negative concept.

\begin{table}
    \centering
    \caption{\textbf{Quantitative evaluation of concept-region alignment using CUB ground-truth segmentation masks.} Positive concept consistently localizes on the object, while unrelated (negative) concept shows near-zero alignment.}
    \label{tab:concept_region_alignment_table}
    \resizebox{\columnwidth}{!}{
    \begin{tabular}{lcc}
    \toprule
    \textbf{Metric} & \textbf{Positive Concept} & \textbf{Negative Concept} \\
    \midrule
    Pointing Accuracy $\uparrow$ & $0.967 \pm 0.180$ & $0.017 \pm 0.128$ \\
    Inside Activation Ratio $\uparrow$ & $0.507 \pm 0.191$ & $0.031 \pm 0.054$ \\
    IoU@10\% $\uparrow$ & $0.423 \pm 0.159$ & $0.019 \pm 0.067$ \\
    IoU@20\% $\uparrow$ & $0.408 \pm 0.148$ & $0.044 \pm 0.087$ \\
    \bottomrule
    \end{tabular}}
\end{table}

\subsection{Time Analysis}
One of the key advantages of \methodname{} is its computational efficiency. Unlike optimization-based decomposition approaches such as SpLiCE~\cite{NEURIPS2024_996bef37} and Z-CBM~\cite{yamaguchi2025zero}, which require solving an optimization problem per image at inference time, our method performs a single linear projection to compute concept activations. Table~\ref{tab:time-analysis} reports the processing time per image for four representative methods evaluated on the ImageNet-100 validation set using a single NVIDIA H100 GPU. While optimization-based methods require iterative solvers at inference time, \methodname{} introduces only a lightweight matrix multiplication ($v_x A$) on top of CLIP's forward pass, keeping inference latency almost identical to CLIP. As shown in Table~\ref{tab:time-analysis}, \methodname{} adds only ${\sim}0.1$\,ms per image over CLIP ($5.90$ vs.\ $5.77$\,ms), whereas Z-CBM and SpLiCE incur $94\times$ and $59\times$ overhead respectively, making \methodname{} suitable for large-scale deployment and interactive analysis.

\begin{table}
    \centering
    \caption{\textbf{Inference time comparison} on ImageNet-100 (NVIDIA H100). We report median per-image latency (ms) with 95\% confidence intervals. \methodname{} adds negligible overhead ($p = 0.31$, Wilcoxon signed-rank test), while SpLiCE and Z-CBM are $59\times$ and $94\times$ slower due to per-image optimization and retrieval.}
    \label{tab:time-analysis}
    \resizebox{\linewidth}{!}{
    \renewcommand{\arraystretch}{1.2}
    \begin{tabular}{lccc}
        \toprule
        \textbf{Method} & \textbf{Embedding (ms/img)} & \textbf{Full Pipeline (ms/img)} & \textbf{Overhead} \\
        \midrule
        CLIP & $0.0001 \pm 0.0000$ & $5.77 \pm 0.55$ & $1.0\times$ \\
        \hline
        Z-CBM & $97.55 \pm 1.33$ & $542.34 \pm 6.02$ & $94.0\times$ \\
        SpLiCE & $4.50 \pm 0.54$ & $338.51 \pm 4.39$ & $58.7\times$ \\
        \rowcolor{blue!8}
        \methodname{} & $0.0006 \pm 0.0000$ & $5.90 \pm 0.73$ & ${\sim}1.0\times$ \\
        \bottomrule
    \end{tabular}
    }
\end{table}

\section{Discussion}
\methodname{} demonstrates that interpretable concept-based reasoning can coexist with strong zero-shot recognition performance. By decomposing CLIP’s vision-language embeddings into a shared concept space, \methodname{} reveals the underlying semantic structure that drives model predictions.  

\myparagraph{Interpretability and Faithfulness.}
\methodname{} attributes each prediction to explicit, human-understandable concepts, offering transparent reasoning paths for both seen and unseen categories. Unlike saliency-based or feature-attribution methods, which provide only localized visual cues, our concept decomposition expresses CLIP’s similarity judgments in terms of global semantic concepts. This enables human inspection of the decision process, as users can identify which concepts most strongly influence each classification.

\myparagraph{Generalization Across Domains.}
The method generalizes effectively across object-centric, fine-grained, and scene-centric datasets. Results on CIFAR-100 and ImageNet-100 show that the concept bottleneck retains nearly all of CLIP’s discriminative power. Meanwhile, performance on CUB and Places365 demonstrates that even in challenging fine-grained or context-heavy domains, concept activations remain semantically aligned. This suggests that CLIP’s internal representations inherently encode human-aligned structures that can be disentangled through our learned concept projection.

\myparagraph{Comparison to Prior Work.}
In contrast to instance-specific methods such as SpLiCE~\cite{NEURIPS2024_996bef37}, which require costly per-sample optimization, \methodname{} learns a single unified concept projection that applies to all data. Compared to retrieval-based approaches such as Z-CBM~\cite{yamaguchi2025zero}, it avoids concept bank search stages, reducing computational overhead. These properties make \methodname{} suitable for large-scale and cross-domain zero-shot applications.

\myparagraph{Limitations and Future Directions.}
For completeness, we describe some limitations of \methodname{}.
First, the linear projection assumption constrains expressive power; highly non-linear semantic relationships may not be fully captured in the concept space. Second, the interpretability depends on the quality and diversity of the concept set, biases in the concept set can affect the fidelity of explanations. Third, the current model focuses on classification; extending the approach to multimodal reasoning tasks remains an open question.  

Future work could explore non-linear concept mappings, adaptive concept discovery, and integration with language models to dynamically expand the concept vocabulary. These directions could further strengthen the interpretability and faithfulness of zero-shot vision-language systems.

\section{Conclusion}
We introduced the \methodname{}, a method that explains CLIP’s zero-shot predictions through human-interpretable concepts. By learning a shared concept projection that aligns image and text embeddings, \methodname{} enables transparent, concept-level reasoning without requiring additional supervision. Through matching and reconstruction objectives, the model preserves CLIP’s semantic structure while exposing the underlying conceptual basis of its predictions.

Across five benchmarks, CIFAR-100, CUB, ImageNet-100, ImageNet-1k, and Places365, \methodname{} maintains strong zero-shot accuracy comparable to CLIP, while providing meaningful explanations at both the image and class levels. Qualitative analyses reveal that the learned concept vectors form coherent semantic clusters, reflecting interpretable structure within CLIP’s embedding space.

Our results demonstrate that open-vocabulary recognition and interpretability need not be mutually exclusive. \methodname{} bridges these paradigms, offering a scalable path toward trustworthy vision-language systems. Future extensions may incorporate adaptive or hierarchical concept spaces, further deepening our understanding of the semantic organization within large-scale multimodal models.

\section*{Acknowledgments}
We acknowledge the computational resources provided by METU Center for Robotics and Artificial Intelligence (METU-ROMER) and TUBITAK ULAKBIM TRUBA. Dr. Alaniz is supported by Hi! PARIS and ANR/France 2030 program (ANR-23-IACL-0005). Dr. Akata acknowledges partial funding by the ERC (853489 - DEXIM) and the Alfried Krupp von Bohlen und Halbach Foundation. Dr. Akbas gratefully acknowledges the support of TUBITAK 2219.

{
    \small
    \bibliographystyle{ieeenat_fullname}
    \bibliography{main}
}

\clearpage
\setcounter{page}{1}
\setcounter{section}{0}
\renewcommand\thesection{\Alph{section}}
\maketitlesupplementary

\section{Implementation Details}
This section provides additional details on concept construction, training procedures, and evaluation settings used for \methodname{}. We omit definitions and equations already introduced in the main paper and focus on implementation-specific clarifications.

\subsection{Backbone and Embedding Extraction}
We primarily use the CLIP RN50 backbone, with additional experiments using ViT-B/32, ViT-L/14 (via OpenAI CLIP), and SigLIP ViT-SO400M/14 (via OpenCLIP). For every dataset, we precompute:

\begin{itemize}[leftmargin=1.4em]
    \item $\ell_2$-normalized image embeddings for all train/validation images,
    \item text embeddings for all dataset class names,
    \item text embeddings for the corresponding concept vocabulary.
\end{itemize}

All embeddings remain frozen during the training of \methodname{}.

\subsection{Concept Vocabulary per Dataset}
We follow the LF-CBM \cite{oikarinenlabel} and use their GPT-3-generated concept sets.

\begin{itemize}[leftmargin=1.4em]
    \item \textbf{ImageNet-1k \& ImageNet-100:}  
    We use the 4{,}751 ImageNet-derived GPT-3 concepts from LF-CBM. No additional concepts are merged.
    \item \textbf{CIFAR-100, CUB, Places365:}  
    The original LF-CBM concept sets for these datasets are limited. Therefore, we merge the dataset's own LF-CBM concepts with the larger ImageNet concept set to obtain a sufficiently expressive concept space. After merging, duplicate concepts are removed.
\end{itemize}

All concepts are encoded once using the corresponding model's text encoder.

\subsection{Training the Concept Projection Matrix $A$}
The learnable matrix $A \in \mathbb{R}^{d \times m}$ maps CLIP/SigLIP embeddings into the $m$-dimensional concept space. The matrix is initialized as
\begin{equation}
    A^{(0)} = \Phi,
\end{equation}
where $\Phi$ is the CLIP concept embedding matrix.

The training objective consists of a matching term and a reconstruction term, as defined in the main paper. The scalar $\lambda$ (typically $\lambda=1$, and $\lambda=5$ for CUB and Places365) controls the relative weight of the reconstruction loss. No orthogonality or sparsity regularizers are used.

We optimize only $A$ (all backbone parameters remain frozen) using Adam with learning rate $10^{-2}$ for 10,000 iterations. After every epoch we renormalize all concept vectors
\begin{equation}
    A_{:,j} \leftarrow \frac{A_{:,j}}{\|A_{:,j}\|_2},
\end{equation}
which stabilizes the concept geometry and prevents drift during training.

\subsection{Zero-Shot and Generalized Zero-Shot Evaluation}
Let $\mathcal{Y}_S$ and $\mathcal{Y}_U$ denote the seen and unseen class sets.

\paragraph{Zero-shot learning (ZSL).}
Evaluation is performed over unseen classes only. For each test image $x$, the predicted label is
\begin{equation}
    \hat{y}
    = \underset{k:\, y_k \in\mathcal{Y}_U}{\arg\max} \; \langle c_x, c_k \rangle,
\end{equation}
and accuracy is computed over the unseen test set $\mathcal{D}_U$ as
\begin{equation}
    \mathrm{Acc}_U =
    \frac{1}{|\mathcal{D}_U|}\sum_{(x,y_k)\in\mathcal{D}_U}
    \mathbbm{1}[\hat{y}=y_k].
\end{equation}

\paragraph{Generalized zero-shot learning (GZSL).}
Predictions are made over the combined label set
$\mathcal{Y}_G = \mathcal{Y}_S \cup \mathcal{Y}_U$.
We report accuracies on seen and unseen classes separately ($\mathrm{Acc}_S$, $\mathrm{Acc}_U$), along with the harmonic mean
\begin{equation}
H = \frac{2 \times \mathrm{Acc}_S \times \mathrm{Acc}_U}{\mathrm{Acc}_S + \mathrm{Acc}_U}.
\end{equation}

\subsection{Time Analysis Protocol}
We measure inference latency on the ImageNet-100 validation set using a single NVIDIA H100 GPU. For each method, we report two quantities: (i) the \emph{embedding time}, which measures only the concept decomposition step (excluding the shared CLIP forward pass), and (ii) the \emph{full pipeline time}, which includes CLIP encoding plus the method-specific decomposition.

To obtain stable estimates, we first run warm-up iterations, then time each method over the full validation set and record per-image latencies. We report the median latency together with 95\% confidence intervals.

For \methodname{}, the decomposition reduces to a single matrix multiplication $c_x = v_x A$, which adds negligible overhead to CLIP's forward pass. To verify this statistically, we apply a Wilcoxon signed-rank test comparing per-image latencies of CLIP and \methodname{}, yielding $p = 0.31$ (no significant difference). In contrast, SpLiCE requires iterative sparse coding per image, and Z-CBM performs a retrieval-based regression over the concept bank, both of which involve iterative optimization and result in substantially higher latency.

\subsection{Concept-Region Alignment}
\label{subsec:suppl_concept_region}

\paragraph{Generating spatial heatmaps.}
To produce concept-level spatial activation maps, we extract patch-level representations from the CLIP RN50 backbone. We register a forward hook on \texttt{layer4} of the ResNet visual encoder to capture the spatial feature map before attention pooling, yielding $N_p = 7 \times 7 = 49$ patch embeddings. Each patch is passed through CLIP's attention pooling projections ($v\_proj$, $c\_proj$) and $\ell_2$-normalized to obtain patch embeddings $\{p_i\}_{i=1}^{N_p}$ in $\mathbb{R}^d$.

Each patch embedding is then projected into the concept space via $A$ to obtain $z_i = p_i A \in \mathbb{R}^m$. We normalize each $z_i$ by its maximum absolute value and mean-center across concepts to ensure comparable activation scales. For a given concept $j$, the spatial heatmap is formed by extracting the $j$-th coordinate of each normalized patch, applying ReLU, and reshaping to the $7 \times 7$ grid. The result is bilinearly upsampled to the original image resolution for visualization.

\paragraph{Quantitative evaluation metrics.}
We evaluate spatial alignment on CUB-200-2011 using ground-truth segmentation masks. For each class, we manually specify a \emph{positive} concept (e.g., \emph{a blue-gray body} for Indigo Bunting) and a \emph{negative} concept (e.g., \emph{a red face}), compute heatmaps for both across all images of that class, and compare against the binary segmentation mask $M$ (resized to the patch grid). We report:

\begin{itemize}[leftmargin=1.4em]
    \item \textbf{Pointing Accuracy}: fraction of images where the maximally activated patch falls inside $M$.
    \item \textbf{Inside Activation Ratio}: proportion of total activation mass falling inside $M$.
    \item \textbf{IoU@$\tau$\%}: intersection-over-union between $M$ and the binary mask obtained by thresholding the heatmap at the $(100{-}\tau)$-th percentile. We report IoU@10\% and IoU@20\%.
\end{itemize}

For each metric, we report the mean and standard deviation across all images in the class. As shown in Table~\ref{tab:concept_region_alignment_table} of the main paper, positive concepts consistently localize on the object region, while negative concepts produce near-zero alignment scores.

\section{Faithfulness \& Causal Validation}
A core requirement for concept-based interpretability is faithfulness: the degree to which the concepts identified as important are causally responsible for the model’s predictions. In this section, we evaluate whether the learned concept space $A$ discovered by \methodname{} produces faithful explanations of CLIP’s zero-shot classifier.

All experiments are conducted on \textbf{ImageNet-100} using the \textbf{CLIP RN50} backbone. Because ImageNet-100 is large, we evaluate faithfulness on a randomly sampled subset of validation images, ensuring stable estimates while keeping computation tractable.

\subsection{Concept Removal for Causal Testing}
As defined in the main paper, the concept-wise interaction score between image $x$ and class $y_k$ is $s_{x,k} = c_x \odot c_k$, where the $j$-th entry $s_{x,k}^{(j)}$ measures how strongly concept $j$ contributes to the prediction. Let $\mathcal{J}_n(x,y_k)$ denote the indices of the top-$n$ concepts ranked by $s_{x,k}^{(j)}$. To test causal influence, we ablate the top-$n$ influential concepts by removing their contribution from the scoring function
\begin{equation}
    f’(x,y_k)
    =
    f(x,y_k)
    -
    \sum_{j \in \mathcal{J}_n(x,y_k)}
    s_{x,k}^{(j)},
\end{equation}
where $f(x,y_k) = \langle c_x, c_k \rangle$ is the concept-space logit reconstructed via \methodname{}.

Faithfulness is quantified as the expected logit drop
\begin{equation}
    \Delta_n
    =
    \mathbb{E}_{(x,y_k)} \big[ f(x,y_k) - f’(x,y_k) \big].
\end{equation}

Higher $\Delta_n$ values imply stronger causal reliance on the discovered concepts.

\subsection{Faithfulness Results}

\paragraph{Mean Logit Drops.}  
Table~\ref{tab:faithfulness} reports the mean logit drop and prediction flip rate after ablating the top-$n$ most influential concepts, averaged across the sampled ImageNet-100 validation images.

\begin{table}[h]
\centering
\caption{\textbf{Faithfulness results on ImageNet-100 (CLIP RN50).} Mean logit drop and prediction flip rate after ablating the top-$n$ most influential concepts.}
\label{tab:faithfulness}
\renewcommand{\arraystretch}{1.15}
\setlength{\tabcolsep}{8pt}
\begin{tabular}{c c c}
\toprule
\textbf{Top-$n$} & \textbf{Logit Drop} & \textbf{Flip Rate} \\
\midrule
1  & 0.0306 & 0.059 \\
3  & 0.0816 & 0.099 \\
5  & 0.1263 & 0.132 \\
10 & 0.2256 & 0.169 \\
\bottomrule
\end{tabular}
\end{table}

Both metrics increase monotonically with $n$: ablating more high-ranked concepts results in larger logit drops and higher flip rates, confirming that the learned concept directions are causally involved in reconstructing CLIP's similarity structure.

\paragraph{Distributional Effects.}

\begin{figure}[h]
    \centering
    \includegraphics[width=0.95\linewidth]{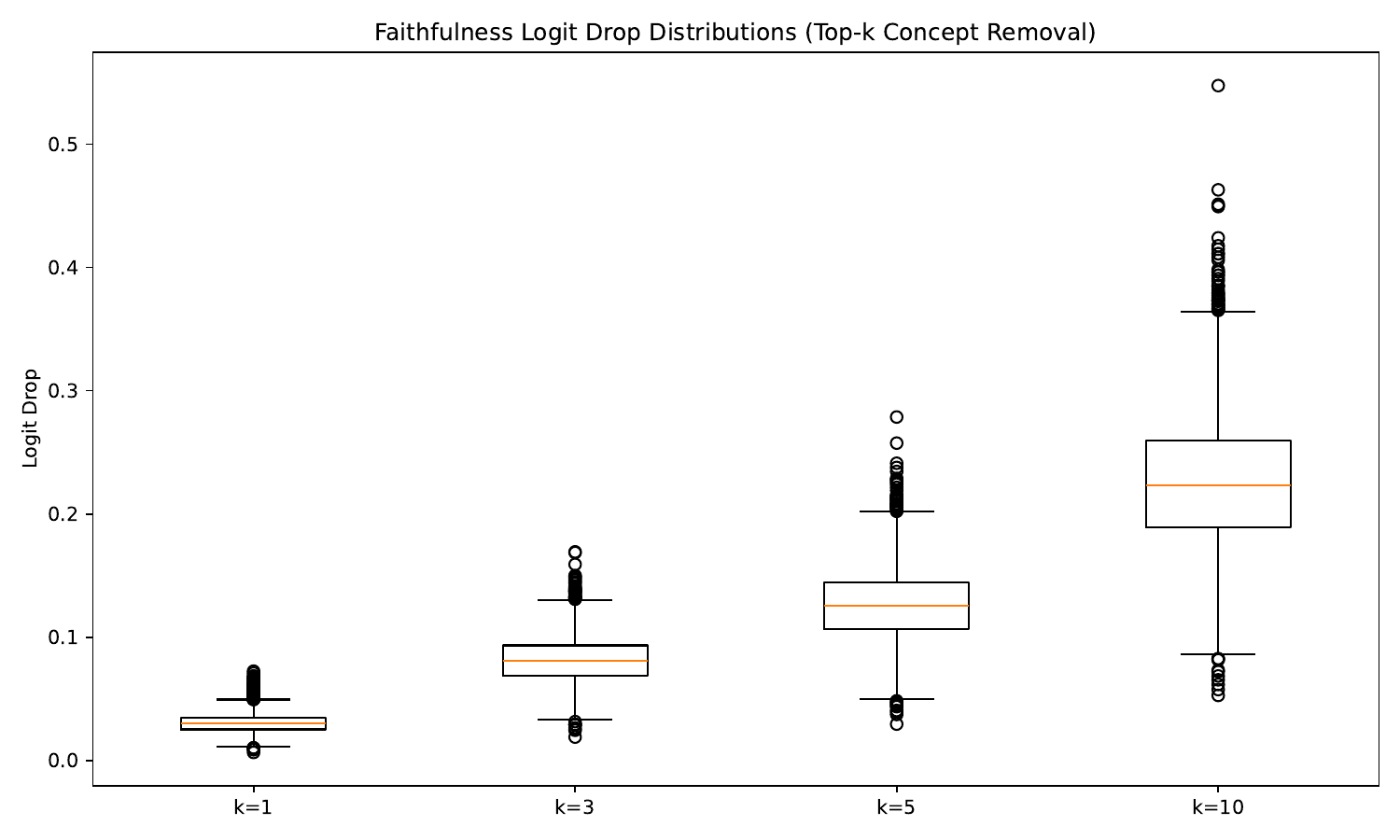}
    \caption{\textbf{Faithfulness distributions for $n{=}1,3,5,10$.} Removing more highly ranked concepts yields consistently larger drops in model confidence.}
    \label{fig:faithfulness_dists}
\end{figure}

Figure~\ref{fig:faithfulness_dists} shows the full empirical distributions of logit drops for $n=\{1,3,5,10\}$.

As $n$ increases, the distributions shift consistently to the right, indicating that the effect is not limited to a few outlier images but holds broadly across the dataset.

\subsection{Causal Intervention: Top-10 vs. Random-10}

To distinguish causal influence from mere correlation, we compare removing the top-10 influential concepts with removing 10 random concepts
\begin{equation}
    \Delta_{\mathrm{top}\text{-}10}
    \quad \text{vs.} \quad
    \Delta_{\mathrm{rand}\text{-}10}.
\end{equation}

Table~\ref{tab:intervention_flip} reports prediction flip statistics when removing the top-10 most influential concepts compared to removing 10 random concepts. Removing the top-10 concepts changes the predicted class for 16.9\% of the evaluated samples, whereas removing 10 random concepts causes flips in only 1.4\%. The large gap between these two settings indicates that the discovered concepts correspond to directions that are causally used by CLIP for decision making rather than reflecting spurious correlations.

\begin{table}[h]
\centering
\caption{\textbf{Top-10 vs Random-10 concept removal.}
Prediction flip counts and flip rates on ImageNet-100 (5000 samples).}
\label{tab:intervention_flip}
\renewcommand{\arraystretch}{1.15}
\setlength{\tabcolsep}{8pt}
\begin{tabular}{l c c}
\toprule
\textbf{Removal type} & \textbf{Flip Count} & \textbf{Flip Rate} \\
\midrule
Top-10 concepts   & 845 & 0.169 \\
Random-10 concepts & 70 & 0.014 \\
\bottomrule
\end{tabular}
\end{table}

\begin{figure}[h]
    \centering
    \includegraphics[width=0.95\linewidth]{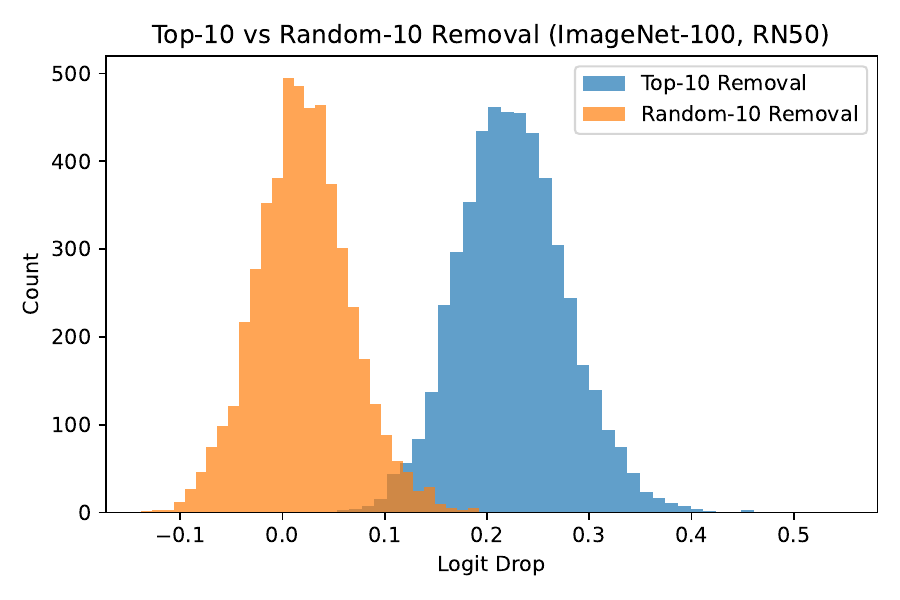}
    \caption{\textbf{Causal intervention analysis.} Removing the top-10 influential concepts produces a substantially larger drop in the predicted class logit than removing 10 random concepts.}
    \label{fig:intervention}
\end{figure}

Figure~\ref{fig:intervention} shows that top-10 removal induces a much larger logit decrease than random removal, whose distribution remains tightly centered near zero. This separation is a strong indicator of true causal involvement: if concept scores merely reflected correlations or dataset priors, random removal would produce similar changes, which it does not.

\subsection{Discussion and Takeaways}

Across all faithfulness and causal tests, the learned concept space $A$ demonstrates:

\begin{itemize}[leftmargin=1.4em]
    \item \textbf{Causal responsibility:}
          Removing top-ranked concepts reliably degrades classifier confidence.
    \item \textbf{Stable, monotonic attribution:}
        Logit drops increase smoothly with $n$, consistent with the additive structure of the concept-space logit.
    \item \textbf{Robustness to dataset variability:}
          Distributions shift consistently, indicating that effects are widespread and not image-specific.
    \item \textbf{No need for sparsity:}
          Despite dense activations, explanations remain faithful. Concept rankings are sufficient for causal attribution.
\end{itemize}

These results confirm that the discovered concept axes form a faithful and causally meaningful basis for explaining CLIP’s zero-shot decisions. The concept directions are not merely descriptive; they directly mediate model predictions in a measurable and controllable manner.

\section{Concept Space Structure Analysis}
This section provides an extended analysis of the geometry and activation behavior of the learned concept space $A$ compared to the original CLIP concept space~$\Phi$. All analyses are conducted on \textbf{ImageNet-100} using the \textbf{CLIP RN50} backbone, matching the quantitative setup of the main paper. Our goal is to understand how optimization alters the concept space, whether semantic identity is preserved, and whether the learned space is geometrically stable, coherent, and suitable for interpretation.

\subsection{How the Concept Spaces Are Obtained}
The original CLIP concept matrix $\Phi$ is created by encoding each concept name $j$ using the CLIP text encoder:
\begin{equation}
\Phi_j = \mathrm{normalize}(f_{\text{text}}(\text{``a photo of }j\text{''})\big).
\end{equation}

The learned concept space $A$ is obtained from our optimization objective, using the same concept vocabulary.

For an image $x$ and class label $y_k$, we compute the image embedding $v_x \in \mathbb{R}^d$ and text embedding $t_k \in \mathbb{R}^d$
\begin{equation}
v_x = \mathrm{normalize}(f_{\text{img}}(x)), \qquad
t_k = \mathrm{normalize}(f_{\text{text}}(y_k)).
\end{equation}

Following the explanation model of the main paper, the image-side, label-side, and joint concept activations are
\begin{equation}
c_x = v_x A,\quad
c_k = t_k A,\quad
s_{x,k} = c_x \odot c_k.
\end{equation}

These activations form the basis for all structure measurements in this section, ensuring full consistency with the interpretability mechanism.

\subsection{Summary of Quantitative Results}

Table~\ref{tab:structure-summary} summarizes the key geometric properties of $A$ and $\Phi$. We provide detailed explanations in the subsections below.

\begin{table}[h]
\centering
\caption{\textbf{Summary of concept space statistics.} All measurements are computed on ImageNet-100 using CLIP RN50. The learned space $A$ preserves semantic identity while becoming more compact and uniformly correlated.}
\label{tab:structure-summary}
\vspace{0.4em}
\scriptsize
\begin{tabular}{lcc}
\toprule
\textbf{Metric} & \textbf{CLIP ($\Phi$)} & \textbf{Learned ($A$)} \\
\midrule
Alignment Mean / Median & - & 0.651 / 0.648\\
Alignment Std & - & 0.036\\
Alignment Min / Max & - & 0.559 / 0.822 \\
Total PCA Variance & 0.1396 & 0.1047 \\
Top-10 Variance Fraction & 0.4586 & 0.4364 \\
Off-diagonal Mean ($\Phi^\top \Phi$ vs. $A^\top A$) & 0.6920 & 0.7461 \\
Off-diagonal Std & 0.0705 & 0.0593 \\
\bottomrule
\end{tabular}
\end{table}

These results support the main claim that our learned concept space remains semantically meaningful and structurally well-behaved.

\subsection{Alignment Between $A$ and $\Phi$}

We evaluate how much each learned concept direction $A_j$ deviates from its original CLIP counterpart $\Phi_j$ using cosine alignment
\begin{equation}
\mathrm{align}(A_j, \Phi_j)
= \left\langle
\frac{A_j}{\|A_j\|},
\frac{\Phi_j}{\|\Phi_j\|}
\right\rangle.
\end{equation}

\begin{figure}[h]
\centering
\includegraphics[width=0.95\linewidth]{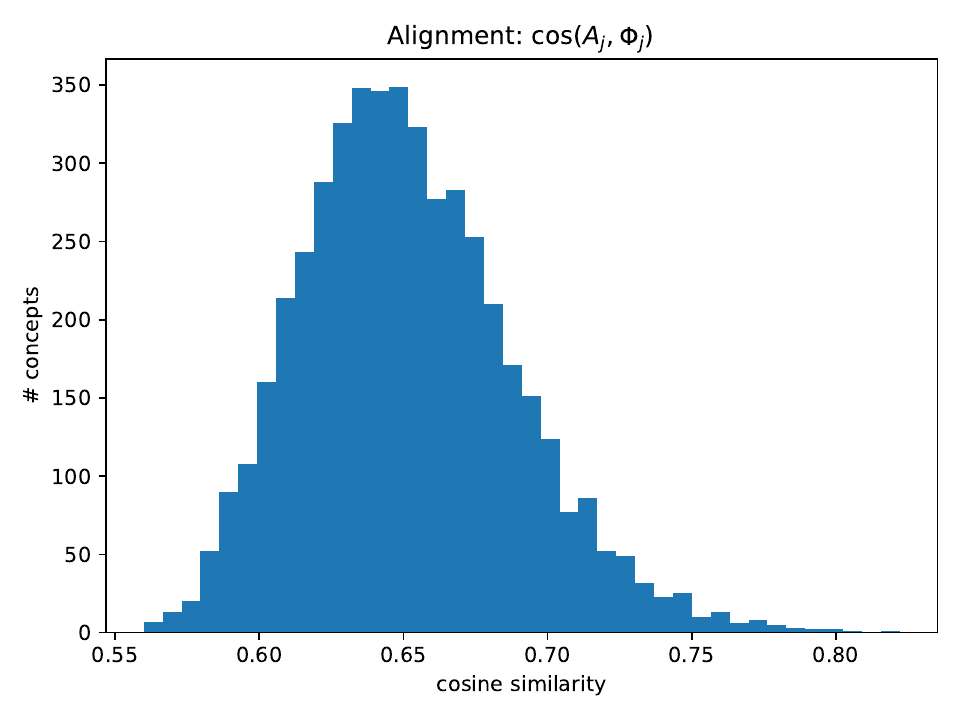}
\caption{\textbf{Alignment distribution} between learned concept directions $A_j$ and original CLIP directions $\Phi_j$. High alignment values indicate semantic consistency between the learned space and CLIP.}
\label{fig:alignment}
\end{figure}

Figure~\ref{fig:alignment} shows that alignment scores are tightly clustered around $0.65$, with a standard deviation of only $0.036$ and minimum/maximum values of $0.559$ and $0.822$. This distribution demonstrates two important facts:

\begin{itemize}[leftmargin=1.4em]
    \item The learned concept directions preserve \textbf{semantic grounding}.
    \item Optimization introduces \textbf{controlled refinements} rather than
          large distortions.
\end{itemize}

Thus, each learned concept remains strongly related to its original semantic meaning, ensuring stable and interpretable explanations.

\subsection{PCA Geometry}

To assess global geometry, we perform PCA on $A$ and $\Phi$, treating concepts as data points in the embedding space.

\begin{figure}[h]
\centering
\includegraphics[width=0.95\linewidth]{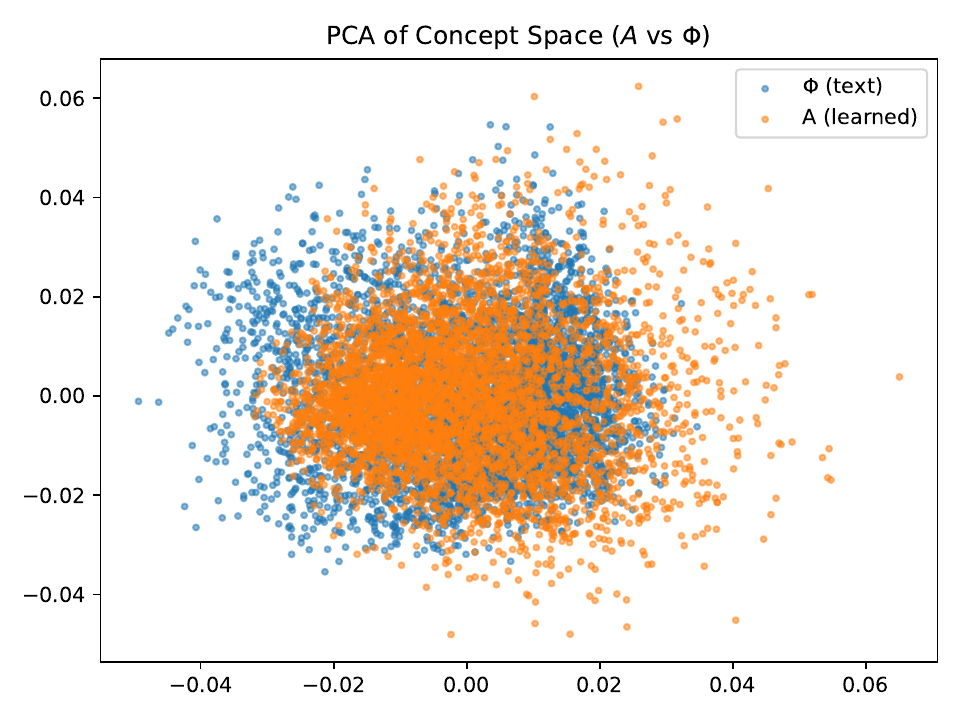}
\caption{\textbf{PCA comparison} between CLIP concept space ($\Phi$) and learned space ($A$). The learned space is more compact yet preserves the structural layout of concept groups.}
\label{fig:pca-A-vs-C}
\end{figure}

\begin{samepage}
The PCA results show:

\begin{itemize}[leftmargin=1.4em]
    \item \textbf{Lower total variance}: $A$ has $0.1047$ vs.\ $0.1396$ in $\Phi$.
    \item \textbf{Comparable top-10 variance fraction}: $0.4364$ vs.\ $0.4586$.
\end{itemize}
\end{samepage}

This means that $A$ is more compact, but not low-rank or collapsed. The PCA scatter in Figure~\ref{fig:pca-A-vs-C} shows that concept clusters and their relative positions are preserved, indicating that the refined space maintains CLIP’s semantic organization while smoothing its geometry.

\subsection{Activation Density and Why Sparsity Is Not Required}

\begin{figure}[h]
\centering
\includegraphics[width=0.95\linewidth]{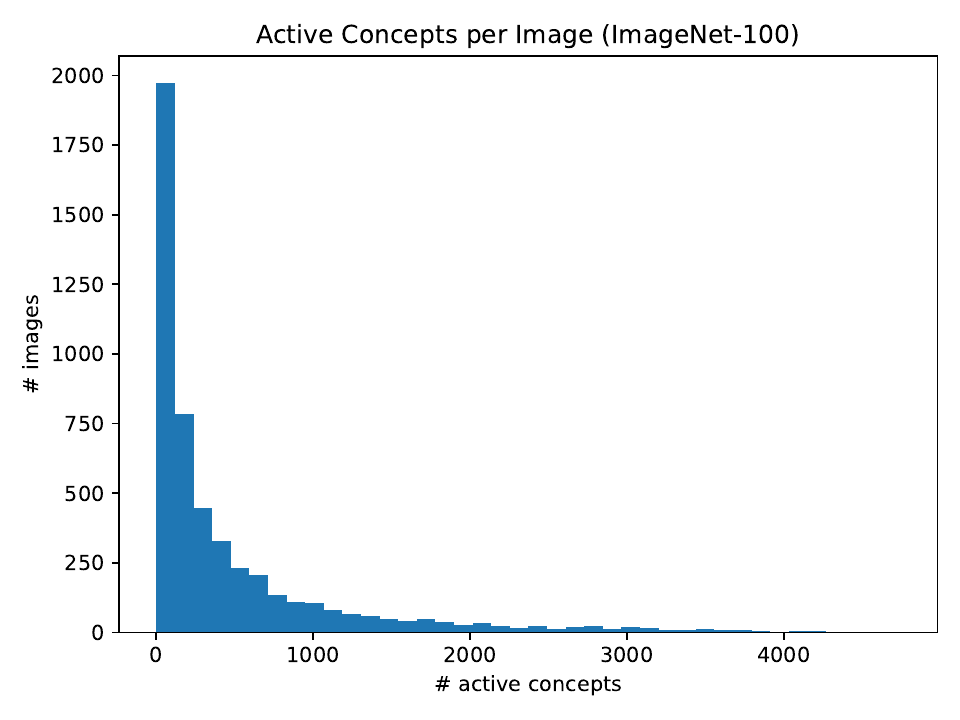}
\caption{\textbf{Activation density}: number of active concepts per image.}
\label{fig:act-img}
\end{figure}

\begin{figure}[h]
\centering
\includegraphics[width=0.95\linewidth]{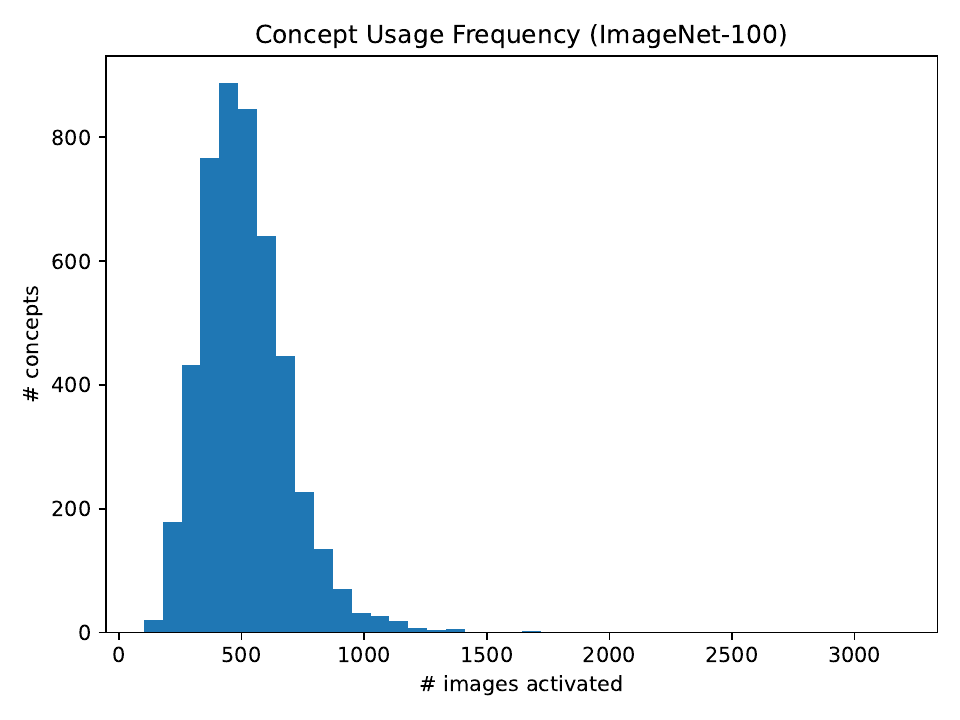}
\caption{\textbf{Activation coverage}: number of images activating each concept.}
\label{fig:act-concept}
\end{figure}

We examine the density of concept activations using the joint interaction vector $s_{x,k} = c_x \odot c_k$. For each image, a concept $j$ is counted as active if $s_{x,k}^{(j)} > \tau$, where $\tau = 0.01$. Figures~\ref{fig:act-img} and \ref{fig:act-concept} show the distribution of active concepts per image and the number of images activating each concept.

\begin{samepage}
Across ImageNet-100:

\begin{itemize}[leftmargin=1.4em]
    \item Each image activates \textbf{490 concepts on average} (median 190).
    \item Each concept is activated for \textbf{516 images on average} (median 494).
\end{itemize}
\end{samepage}

These dense activation patterns are expected because CLIP embeddings are intrinsically distributed. Importantly, \textbf{sparsity is not required} for concept-based interpretability in our method. Explanations rely on identifying and ranking the most influential concept directions (top-$n$), not on enforcing a few active dimensions. Dense signals still yield clear, semantically aligned top concepts, as demonstrated in the qualitative results of the main paper.

\section{CLIP-EZPC Fidelity}
We evaluate how closely EZPC reproduces the predictions of the original CLIP model. This measures the faithfulness of the learned concept space to the teacher model. We report the following metrics: \emph{Top-1 agreement}, \emph{Spearman rank correlation}, \emph{Kendall rank correlation}, and \emph{KL divergence}. Results are shown in Table~\ref{tab:fidelity_metrics}.

\begin{table}
    \centering
    \caption{\textbf{Prediction-level fidelity between CLIP and EZPC (RN50 backbone) with label consistency, ranking preservation, and KL.} Top-1 agreement measures label consistency. Spearman and Kendall correlations quantify ranking preservation. Kendall correlation is computed over the top-50 CLIP-ranked classes.}
    \label{tab:fidelity_metrics}
    \resizebox{\columnwidth}{!}{
    \begin{tabular}{lcccc}
    \toprule
    Dataset & Top-1 Agree. (\%) & Spearman & Kendall & KL \\
    \midrule
    CIFAR-100     & 80.26 & 0.959 & 0.733 & $6.79 \times 10^{-6}$ \\
    IN-100  & 92.92 & 0.994 & 0.904 & $6.10 \times 10^{-6}$ \\
    CUB           & 84.03 & 0.994 & 0.876 & $6.83 \times 10^{-6}$ \\
    ImageNet-1k   & 72.37 & 0.924 & 0.536 & $7.98 \times 10^{-5}$ \\
    Places365     & 79.92 & 0.972 & 0.715 & $1.46 \times 10^{-5}$ \\
    \bottomrule
    \end{tabular}}
\end{table}

Fidelity remains high across datasets, indicating that EZPC preserves the ranking structure of CLIP predictions while enabling concept-based explanations. Lower agreement on larger datasets, such as ImageNet-1k, reflects the increased semantic ambiguity rather than the failure of the concept model.

\section{Additional Ablation Studies}
\subsection{Impact of Concept Vocabulary Size}

We analyze how the number of concepts affects both predictive performance and explanation fidelity. We randomly subsample the concept vocabulary and train EZPC with $m \in \{250, 500, 1000, 2000, 3000, 4751\}$ concepts. For each size, we repeat training with three random seeds. Results are shown in Table~\ref{tab:vocab_size_ablation_table}.

\begin{table}[h]
    \centering
    \caption{\textbf{Effect of concept vocabulary size $m$ on performance, tested on IN-100 under the generalized ZSL setting using the RN50 backbone.}}
    \label{tab:vocab_size_ablation_table}
    \resizebox{\columnwidth}{!}{
    \begin{tabular}{lcccc}
    \toprule
    $m$ & Top-1 Agree. (\%) & Seen Acc. & Unseen Acc. & H \\
    \midrule
    250  & $69.05 \pm 0.34$ & $0.5527 \pm 0.0027$ & $0.5637 \pm 0.0105$ & $0.5581 \pm 0.0054$ \\
    500  & $80.95 \pm 1.35$ & $0.6252 \pm 0.0064$ & $0.6423 \pm 0.0090$ & $0.6336 \pm 0.0072$ \\
    1000 & $89.09 \pm 0.33$ & $0.6622 \pm 0.0028$ & $0.6827 \pm 0.0032$ & $0.6723 \pm 0.0030$ \\
    2000 & $91.73 \pm 0.17$ & $0.6702 \pm 0.0017$ & $0.6890 \pm 0.0026$ & $0.6795 \pm 0.0004$ \\
    3000 & $92.58 \pm 0.26$ & $0.6742 \pm 0.0010$ & $0.6900 \pm 0.0026$ & $0.6820 \pm 0.0010$ \\
    4751 (full) & $92.92$ & $0.6745$ & $0.6900$ & $0.6821$ \\
    \bottomrule
    \end{tabular}}
\end{table}

Performance improves monotonically with the number of concepts, with diminishing returns beyond approximately 3000 concepts. This shows that EZPC is robust to moderate reductions in vocabulary size, while larger vocabularies mainly improve prediction fidelity.

\subsection{Impact of Training Objectives}
We analyze the contribution of each component of the training objective used to learn the projection matrix $A$. Our method optimizes a combination of a matching loss $\mathcal{L}_{\text{match}}$ and a reconstruction loss $\mathcal{L}_{\text{recon}}$. 

\begin{samepage}
\noindent We compare the following settings:

\begin{itemize}
    \item No training ($A = \Phi$)
    \item Matching loss only
    \item Reconstruction loss only
    \item Full objective ($\mathcal{L}_{\text{match}} + \lambda \mathcal{L}_{\text{recon}}$)
\end{itemize}
\end{samepage}

Results calculated on ImageNet-100 using the generalized zero-shot setting are shown in Table~\ref{tab:loss_ablation}.

\begin{table}[h]
    \centering
    \caption{\textbf{Ablation study on the training objective of projection matrix $A$ using RN50 on IN-100 (generalized zero-shot setting).}}
    \label{tab:loss_ablation}
    \resizebox{\columnwidth}{!}{
    \begin{tabular}{lccc}
    \toprule
    Training Setting & Seen Acc. & Unseen Acc. & H \\
    \midrule
    $A = \Phi$ (0-step, no training) & 0.013 & 0.0 & 0.0 \\
    Matching loss only ($\mathcal{L}_{\text{match}}$) & 0.013 & 0.0 & 0.0 \\
    Reconstruction loss only ($\mathcal{L}_{\text{recon}}$) & 0.680 & 0.708 & 0.693 \\
    Full objective ($\mathcal{L}_{\text{match}} + \lambda \mathcal{L}_{\text{recon}}$) & 0.674 & 0.690 & 0.682 \\
    \bottomrule
    \end{tabular}}
\end{table}

Without any training ($A = \Phi$) or with the matching loss alone, the model achieves near-zero accuracy. This is expected as the raw CLIP concept embeddings do not form a basis that can reconstruct the original similarity structure, and the matching loss by itself only regularizes $A$ toward $\Phi$ without learning to preserve predictive information. The reconstruction loss alone achieves the highest harmonic mean ($H = 0.693$), indicating that it is the primary driver of classification performance. Adding the matching loss in the full objective slightly reduces the harmonic mean to $0.682$, but as shown in Figure~\ref{fig:lambda_comparison}, the matching loss plays a critical role in maintaining concept interpretability. Without it, learned concept directions drift from their original semantic meaning, producing less interpretable explanations despite higher quantitative scores. The full objective, therefore, represents a trade-off between predictive fidelity and explanation quality.

\section{Additional Qualitative Visualizations}
\begin{figure*}
  \centering
  \includegraphics[width=\textwidth]{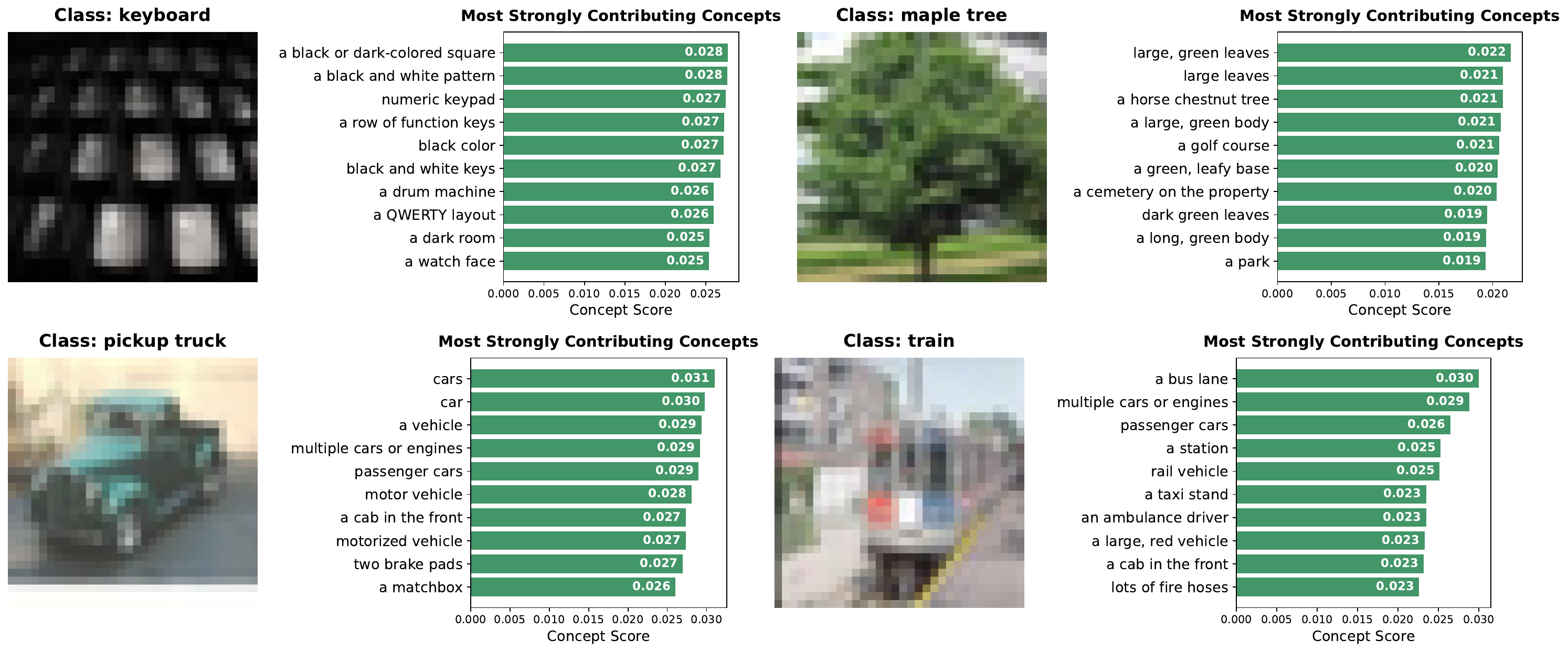}
  \caption{CIFAR-100 image-level explanations.}
  \label{fig:cifar100_image_level}
\end{figure*}

\begin{figure*}
  \centering
  \includegraphics[width=\textwidth]{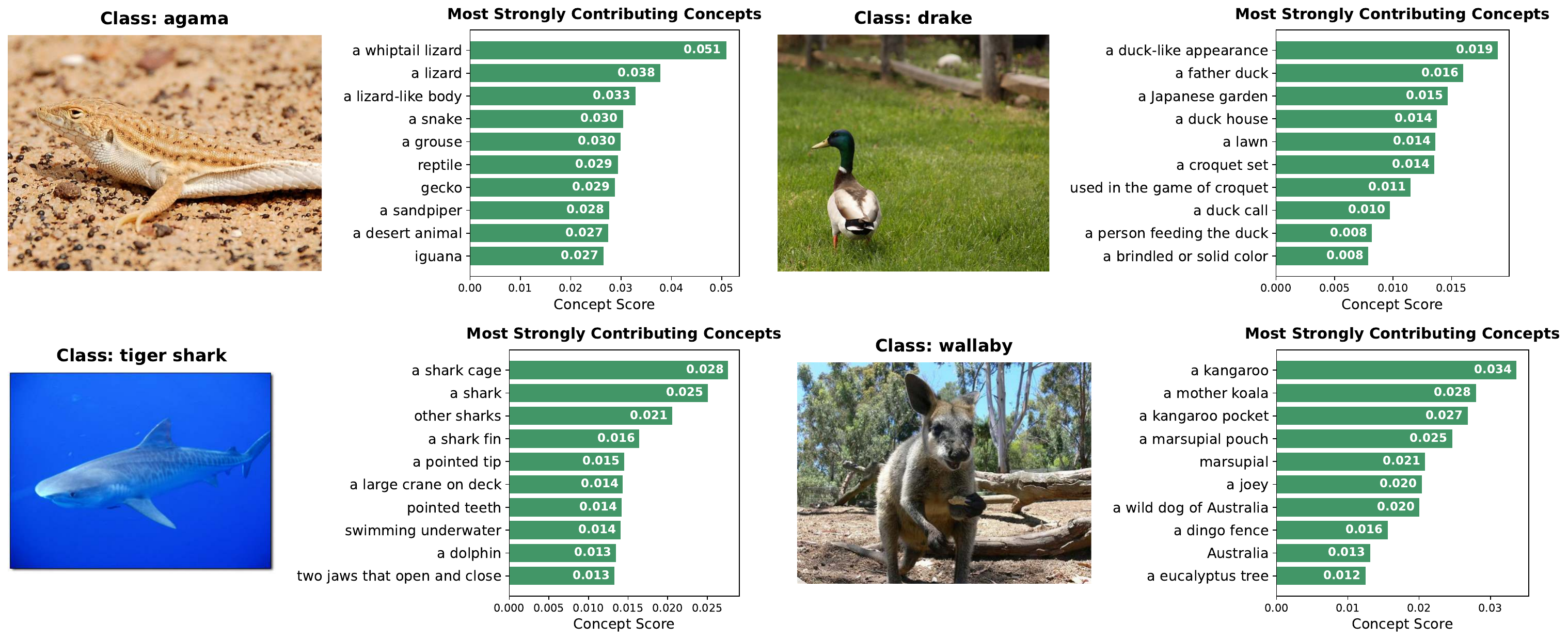}
  \caption{ImageNet-100 image-level explanations.}
  \label{fig:imagenet100_image_level}
\end{figure*}

\begin{figure*}
  \centering
  \includegraphics[width=\textwidth]{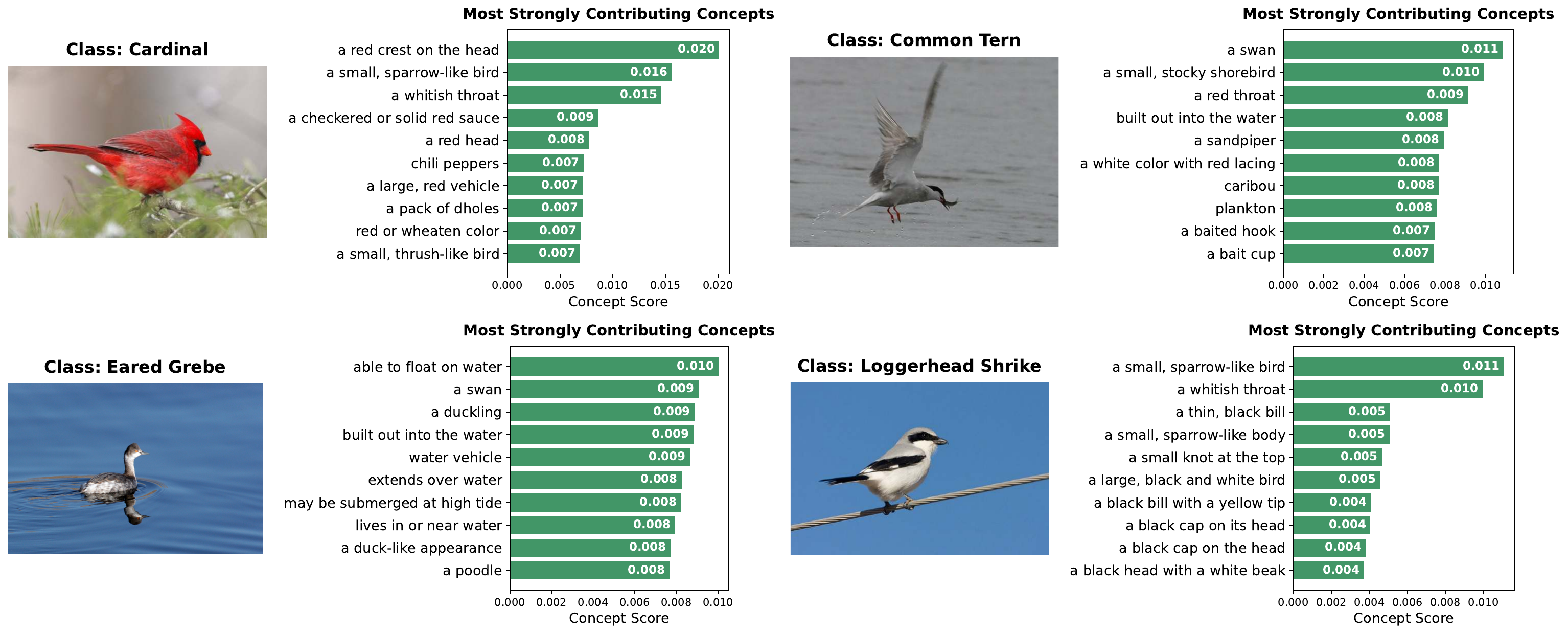}
  \caption{CUB image-level explanations.}
  \label{fig:cub_image_level}
\end{figure*}

\begin{figure*}
  \centering
  \includegraphics[width=\textwidth]{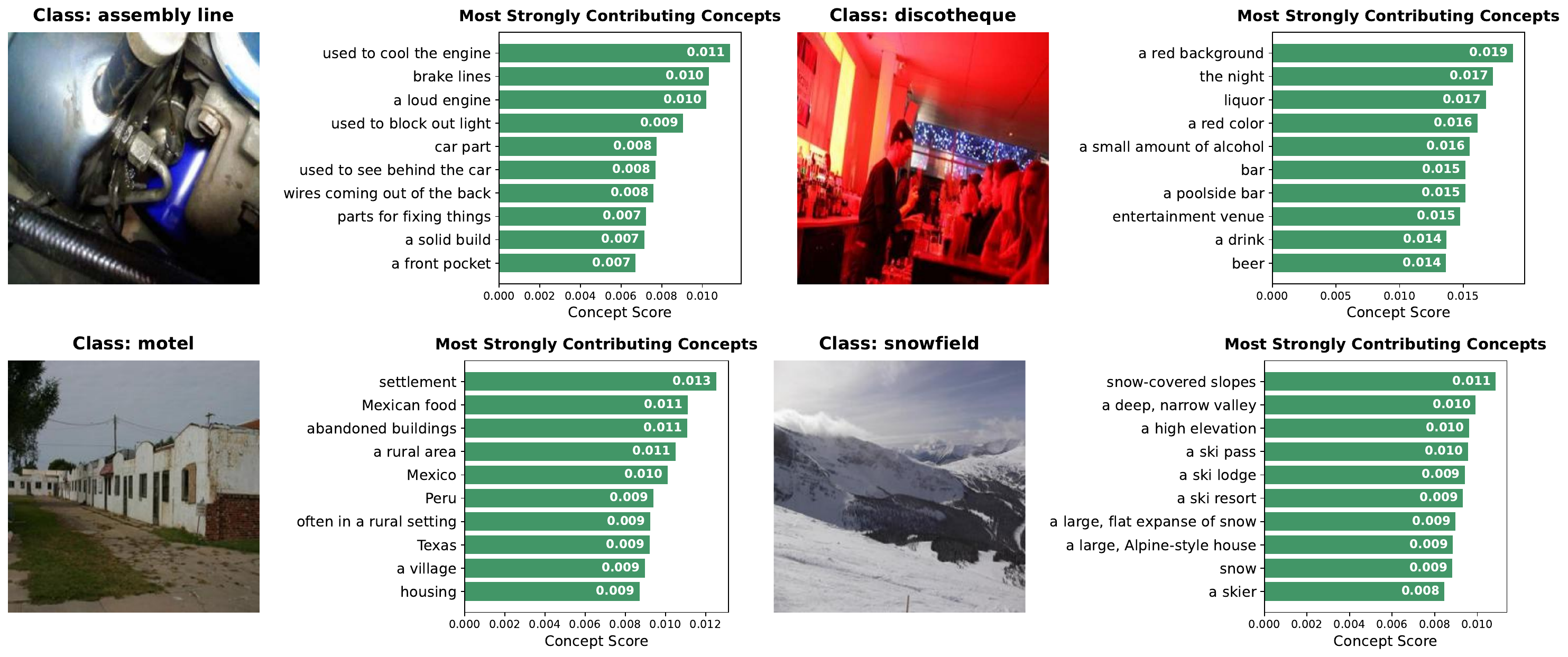}
  \caption{Places365 image-level explanations.}
  \label{fig:places365_image_level}
\end{figure*}

\begin{figure*}
  \centering
  \includegraphics[width=\textwidth]{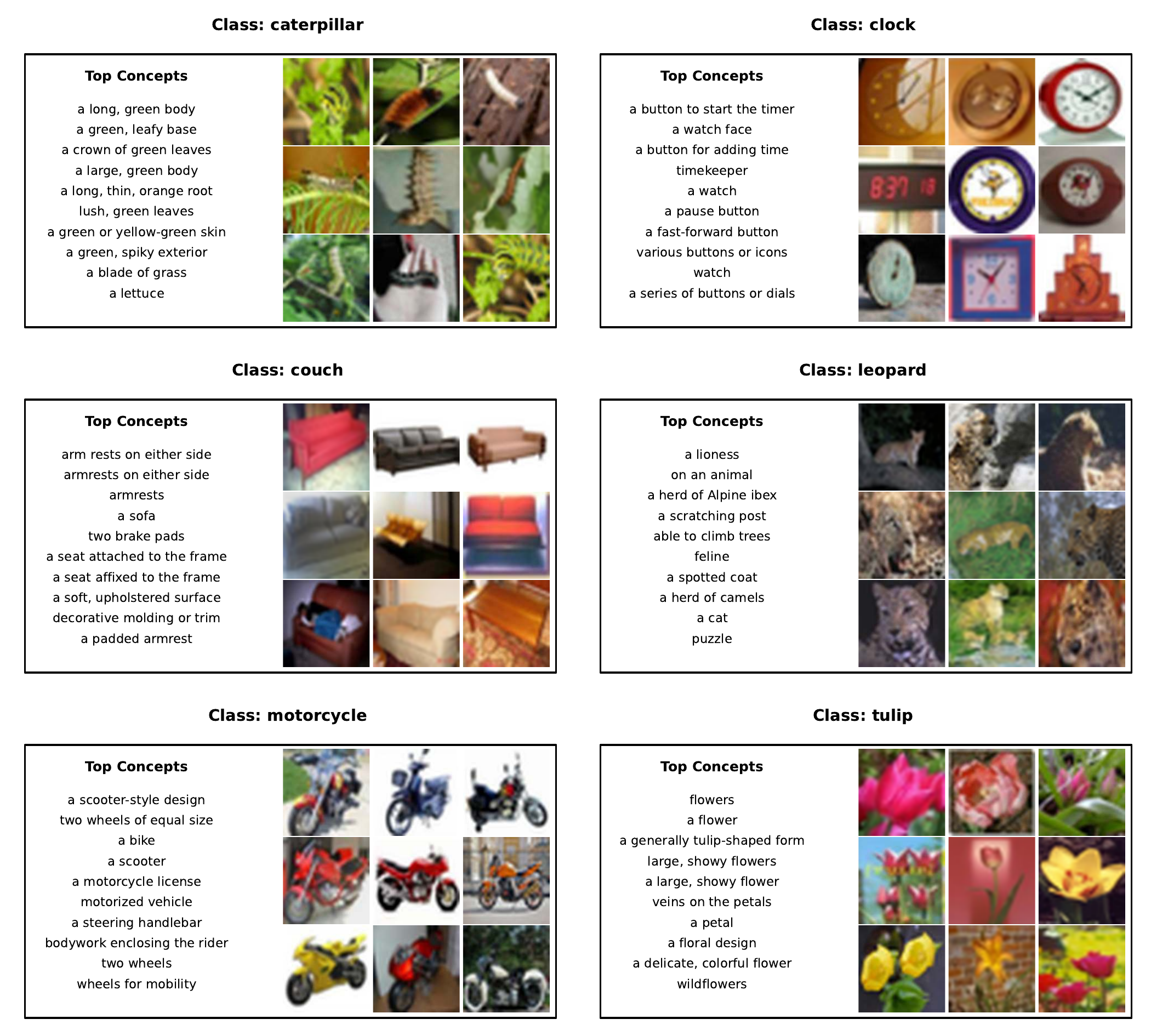}
  \caption{CIFAR-100 class-level visualizations.}
  \label{fig:cifar100_class_level}
\end{figure*}

\begin{figure*}
  \centering
  \includegraphics[width=\textwidth]{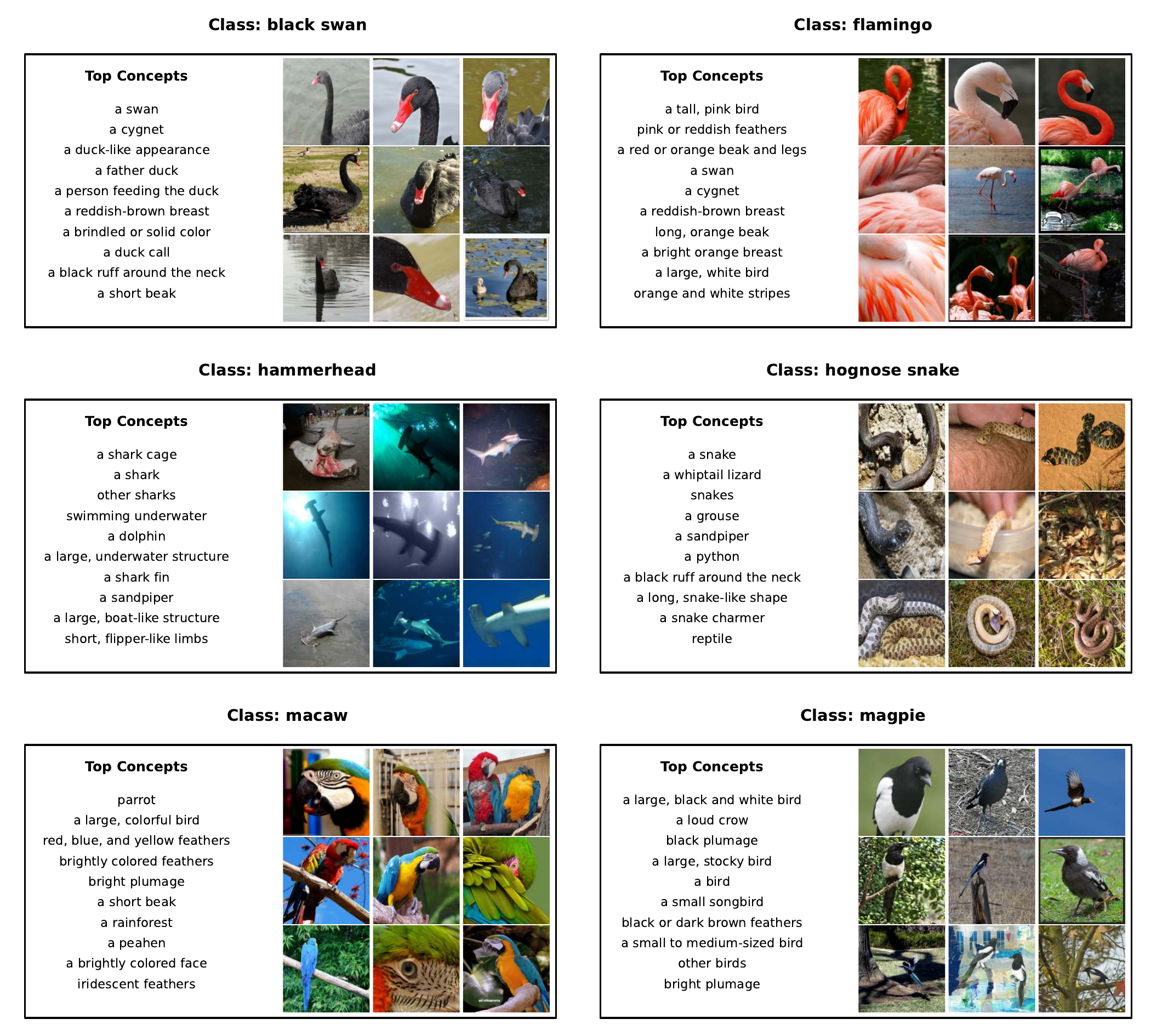}
  \caption{ImageNet-100 class-level visualizations.}
  \label{fig:imagenet100_class_level}
\end{figure*}

\begin{figure*}
  \centering
  \includegraphics[width=\textwidth]{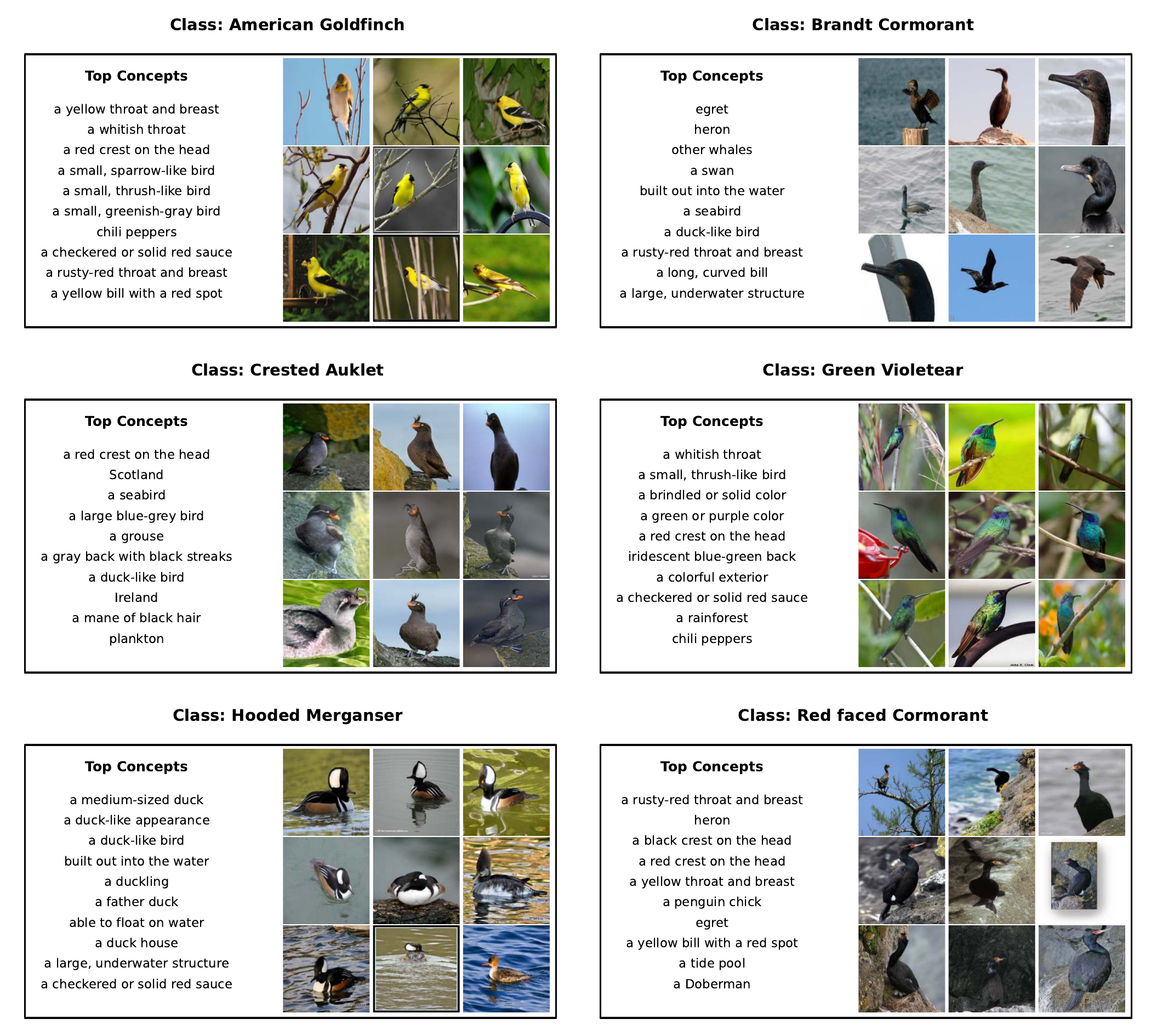}
  \caption{CUB class-level visualizations.}
  \label{fig:cub_class_level}
\end{figure*}

\begin{figure*}
  \centering
  \includegraphics[width=\textwidth]{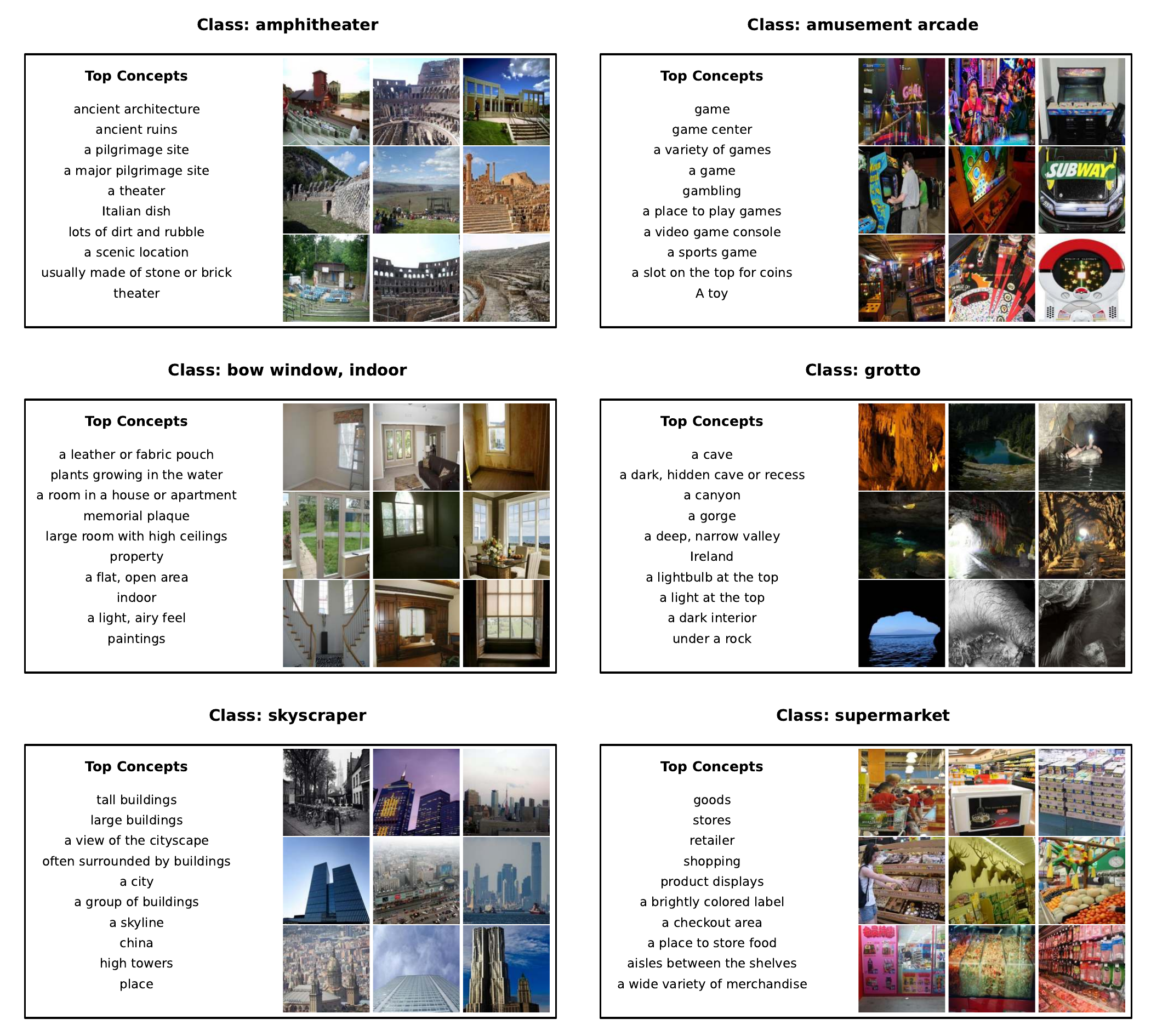}
  \caption{Places365 class-level visualizations.}
  \label{fig:places365_class_level}
\end{figure*}

\begin{figure*}
  \centering
  \begin{subfigure}{\textwidth}
    \centering
    \includegraphics[width=0.77\textwidth]{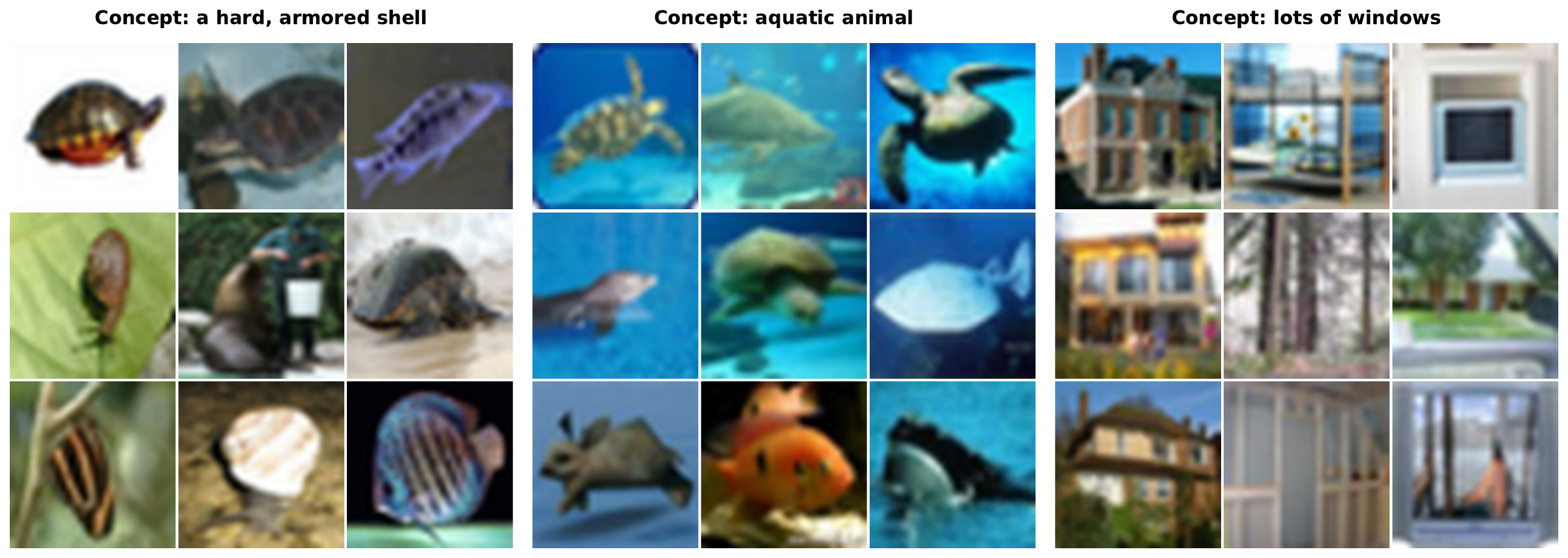}
    \caption{CIFAR-100 concept clustering examples.}
    \label{fig:cifar100_clustering}
  \end{subfigure}
  
  \vspace{1em}
  
  \begin{subfigure}{\textwidth}
    \centering
    \includegraphics[width=0.77\textwidth]{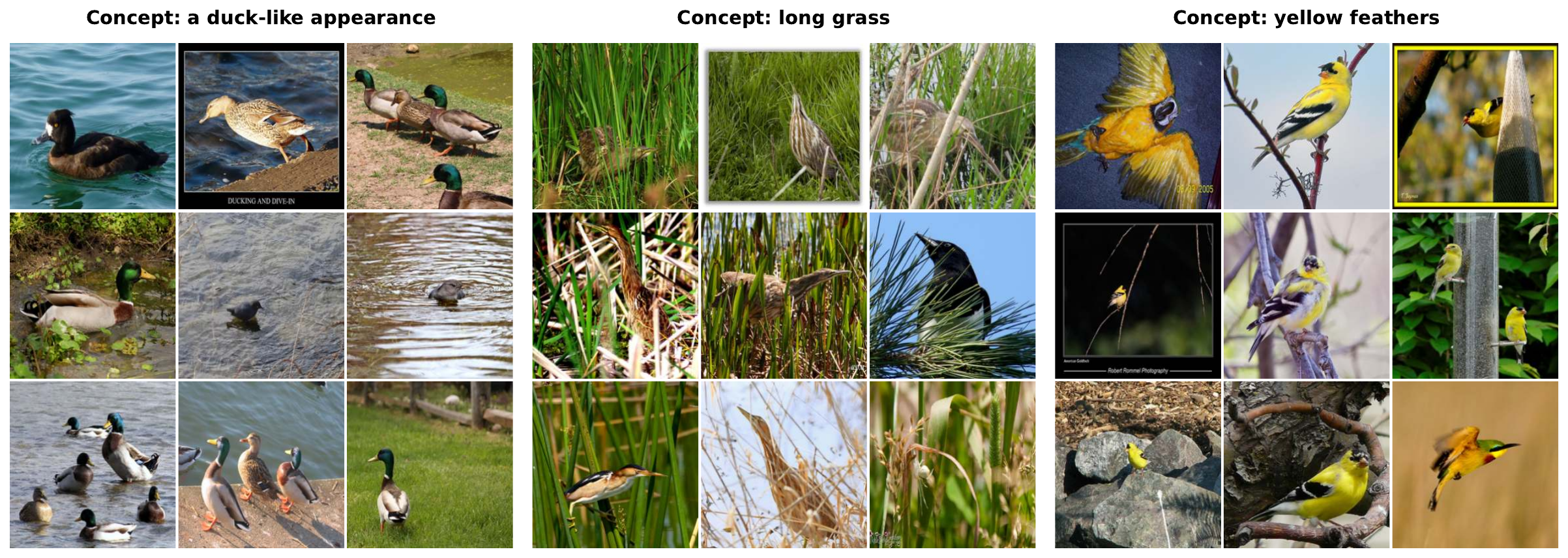}
    \caption{ImageNet-100 concept clustering examples.}
    \label{fig:imagenet100_clustering}
  \end{subfigure}
  
  \vspace{1em}
  
  \begin{subfigure}{\textwidth}
    \centering
    \includegraphics[width=0.77\textwidth]{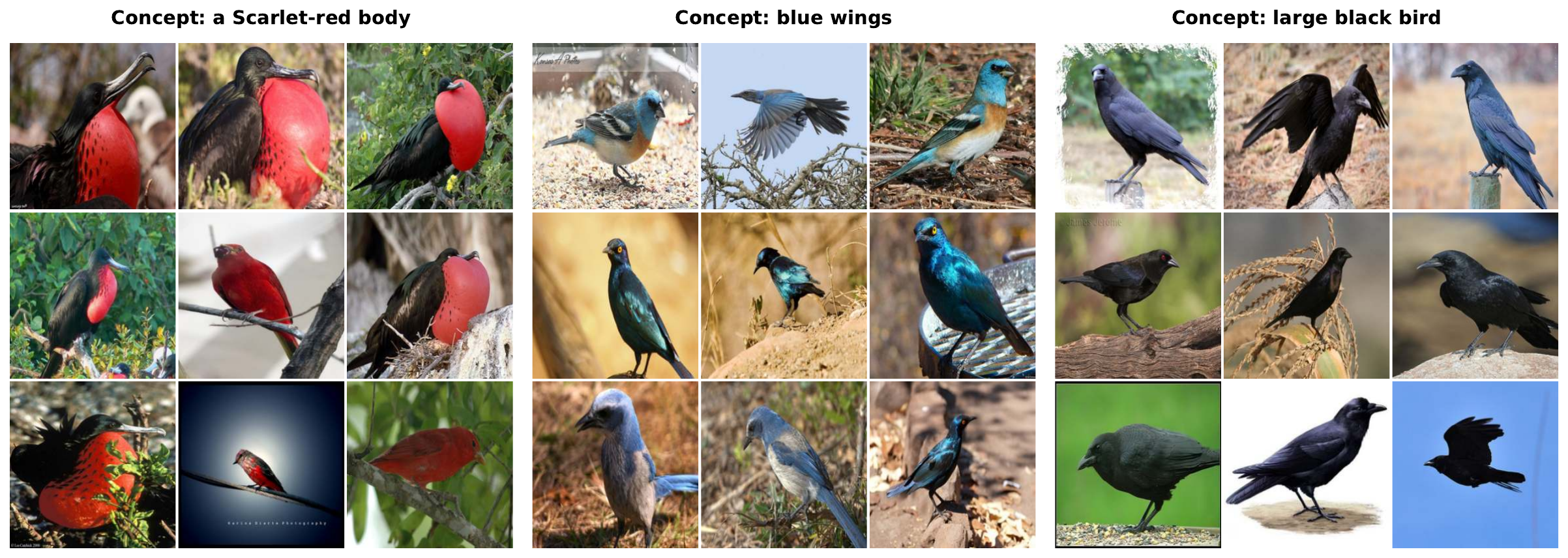}
    \caption{CUB concept clustering examples.}
    \label{fig:cub_clustering}
  \end{subfigure}
  
  \vspace{1em}
  
  \begin{subfigure}{\textwidth}
    \centering
    \includegraphics[width=0.77\textwidth]{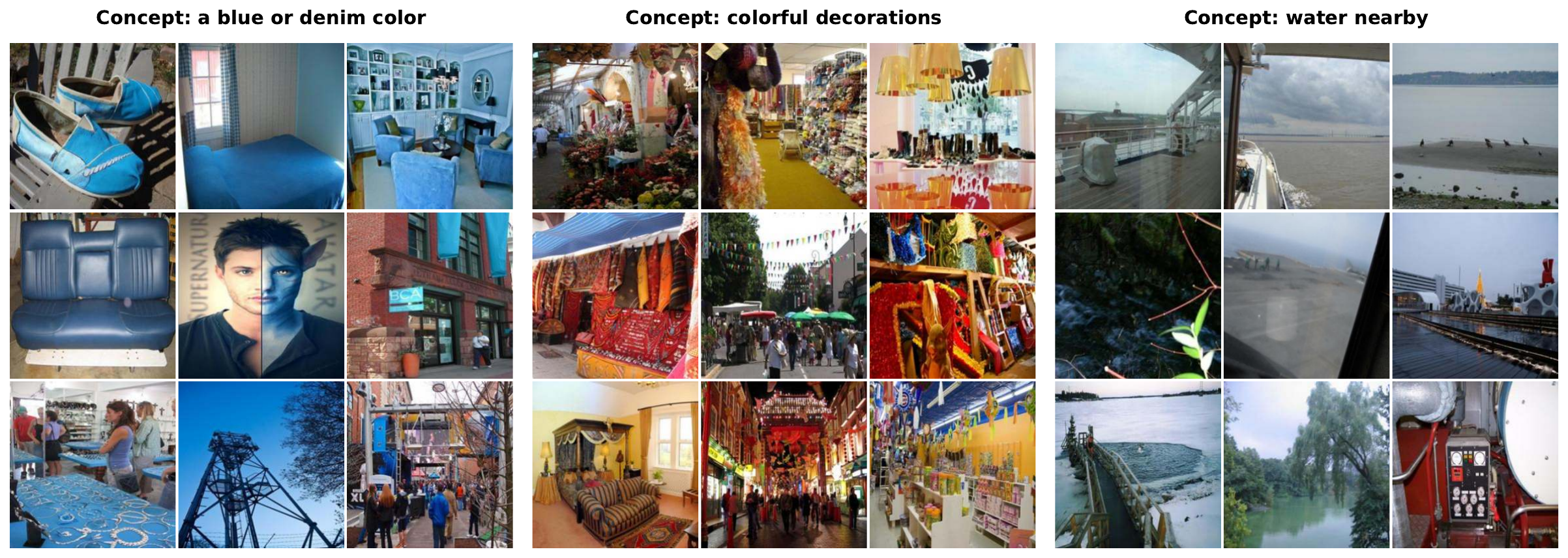}
    \caption{Places365 concept clustering examples.}
    \label{fig:places365_clustering}
  \end{subfigure}
  
  \caption{Concept clustering examples across datasets.}
  \label{fig:all_clustering}
\end{figure*}

\begin{figure*}
  \centering
  \begin{subfigure}{\textwidth}
    \centering
    \includegraphics[width=0.6\textwidth]{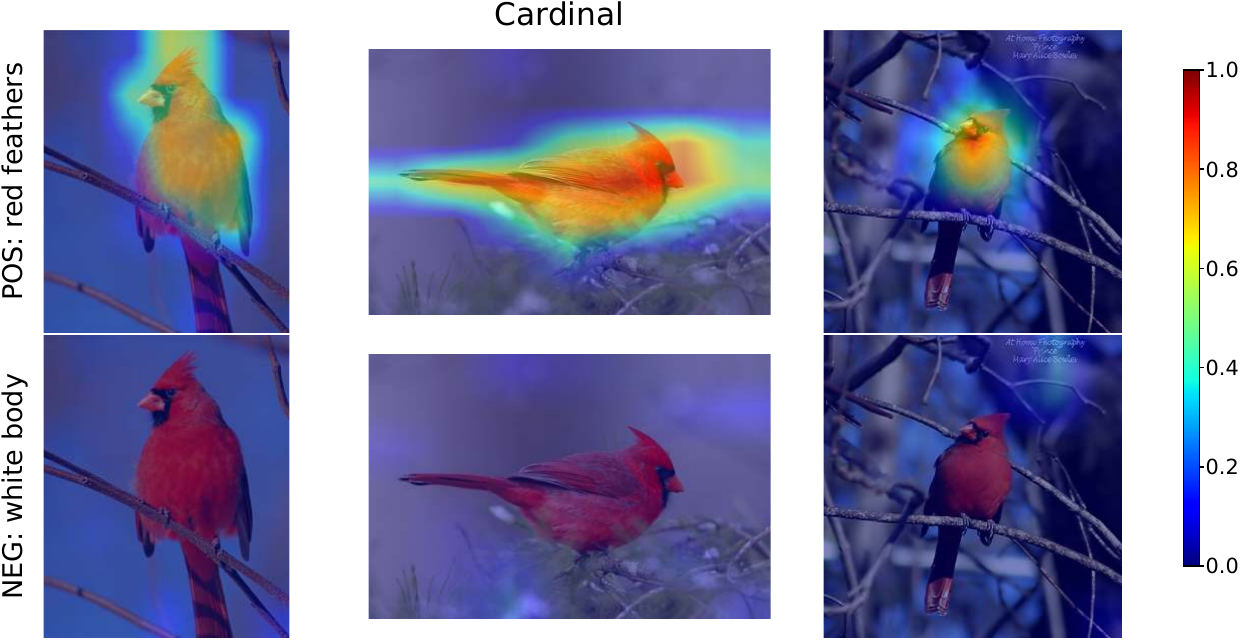}
    \caption{\textbf{Region-level concept alignment for Cardinal class from CUB dataset.} Top row: positive concept (\emph{red feathers}). Bottom row: negative concept (\emph{white body}).}
    \label{fig:cardinal_region_alignment}
  \end{subfigure}

  \vspace{1em}

  \begin{subfigure}{\textwidth}
    \centering
    \includegraphics[width=0.6\textwidth]{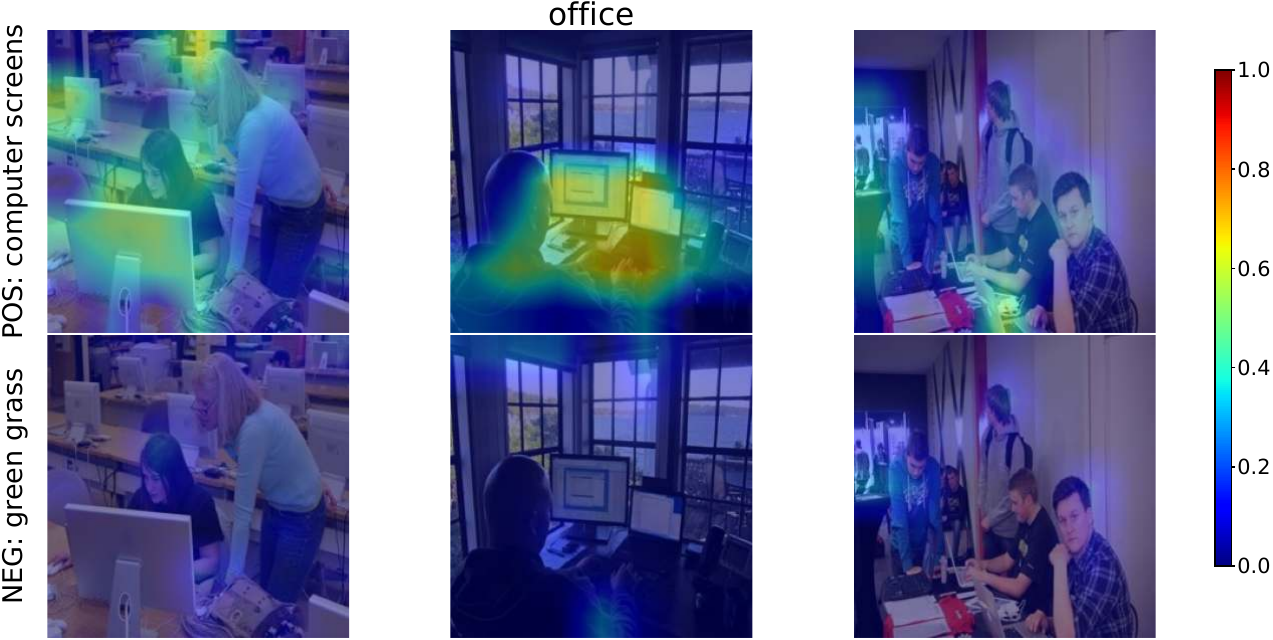}
    \caption{\textbf{Region-level concept alignment for Office class from Places365 dataset.} Top row: positive concept (\emph{computer screens}). Bottom row: negative concept (\emph{green grass}).}
    \label{fig:office_region_alignment}
  \end{subfigure}

  \vspace{1em}

  \begin{subfigure}{\textwidth}
    \centering
    \includegraphics[width=0.6\textwidth]{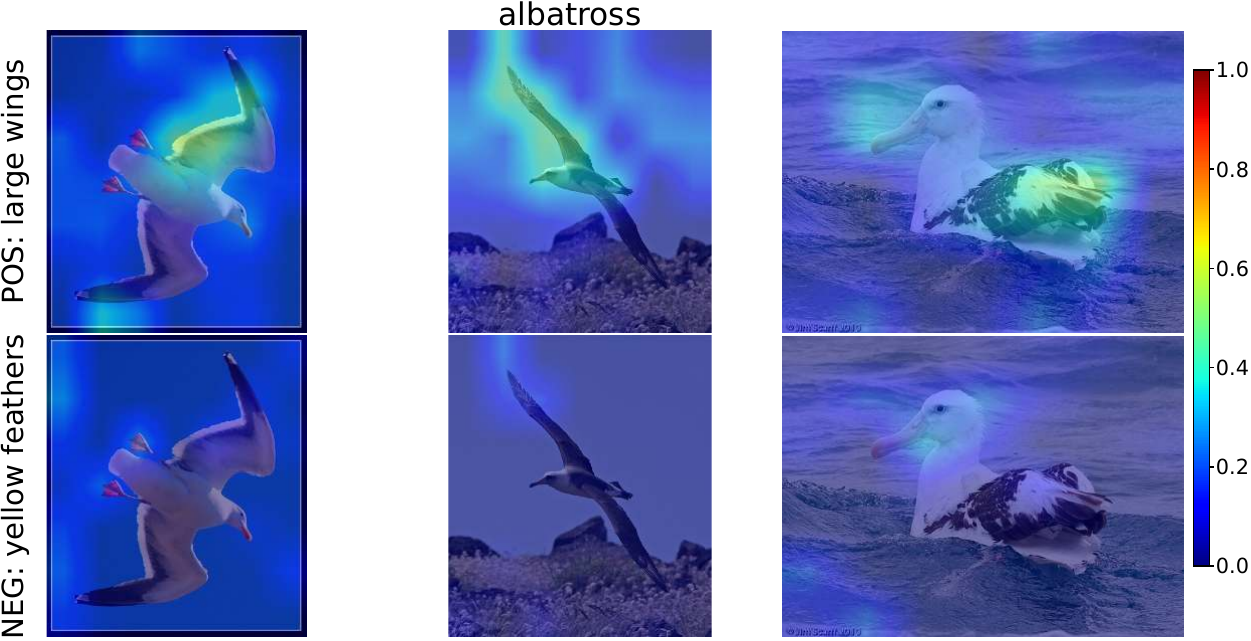}
    \caption{\textbf{Region-level concept alignment for Albatross class from ImageNet-100 dataset.} Top row: positive concept (\emph{large wings}). Bottom row: negative concept (\emph{yellow feathers}).}
    \label{fig:albatross_region_alignment}
  \end{subfigure}

  \caption{Region-level concept alignment examples across datasets.}
  \label{fig:all_region_alignment}
\end{figure*}

This section presents additional qualitative results that complement the analysis in the main paper, covering image-level, class-level, concept-clustering, and region-level alignment visualizations:

\vspace{1em}

\begin{itemize}[leftmargin=1.4em]
    \item \textbf{Image-level explanations (Figures \ref{fig:cifar100_image_level} to \ref{fig:places365_image_level})} Top activated concepts for individual predictions.

    \vspace{0.5em}
    
    \item \textbf{Class-level explanations (Figures \ref{fig:cifar100_class_level} to \ref{fig:places365_class_level})} Average concept activations across a random subset of images per class.

    \vspace{0.5em}
    
    \item \textbf{Concept clustering (Figures \ref{fig:cifar100_clustering} to \ref{fig:places365_clustering})} Clusters of images that strongly activate the given concept.

    \vspace{0.5em}
    
    \item \textbf{Region-level concept alignment (Figures \ref{fig:cardinal_region_alignment} to \ref{fig:albatross_region_alignment})} Spatial localization of positive and negative concepts within individual images.
\end{itemize}

\end{document}